\documentclass[10pt,letterpaper]{article}
\usepackage[top=0.85in,left=2.75in,footskip=0.75in]{geometry}

\usepackage{amsmath,amssymb}

\usepackage{changepage}

\usepackage[utf8x]{inputenc}

\usepackage{textcomp,marvosym}

\usepackage{cite}

\usepackage{nameref,hyperref}

\usepackage[right]{lineno}

\usepackage{microtype}
\DisableLigatures[f]{encoding = *, family = * }

\usepackage[table]{xcolor}

\usepackage{array}

\usepackage{booktabs}

\newcolumntype{+}{!{\vrule width 2pt}}

\newlength\savedwidth

\raggedright
\usepackage[aboveskip=1pt,labelfont=bf,labelsep=period,justification=raggedright,singlelinecheck=off]{caption}

\makeatletter
\renewcommand{\@biblabel}[1]{\quad#1.}
\makeatother

\usepackage{lastpage,fancyhdr,graphicx}
\usepackage{epstopdf}
\fancyheadoffset[L]{2.25in}
\fancyfootoffset[L]{2.25in}
\let\citep\cite
\let\citet\cite

\definecolor{bleudefrance}{rgb}{0.19, 0.55, 0.91}

\renewenvironment{abstract}{\section*{Abstract}}{}

\usepackage{cleveref}
\usepackage{multirow}

\begin{document}
\vspace*{0.2in}

\begin{flushleft}
{\Large
\textbf\newline{Quantifying Rodda and Graham Gait Classification from 3D Markerless Kinematics derived from a Single-view Video in a Heterogeneous Pediatric Clinical Cohort}
}
\newline
\\
Lauhitya Reddy\textsuperscript{1},
Seth Donahue\textsuperscript{2},
Jeremy Bauer\textsuperscript{2},
Susan Sienko\textsuperscript{2},
Anita Bagley\textsuperscript{2},
Joseph Krzak\textsuperscript{2},
Maura Eveld\textsuperscript{2},
Karen Kruger\textsuperscript{2},
Ross Chafetz\textsuperscript{2},
Vedant Kulkarni\textsuperscript{2},
Hyeokhyen Kwon\textsuperscript{1,3*}
\\
\bigskip
\textbf{1} Department of Biomedical Informatics, Emory University, Atlanta, GA, USA
\\
\textbf{2} Shriners Children's, USA
\\
\textbf{3} The Wallace H. Coulter Department of Biomedical Engineering, Emory University and Georgia Institute of Technology, Atlanta, GA, USA
\\
\bigskip

* hyeokhyen.kwon@emory.edu

\end{flushleft}

\begin{abstract}
Cerebral Palsy (CP) is a non-progressive neurological disorder of movement and the most common cause of lifelong physical disability in childhood.
Approximately 75\% of children with CP are ambulatory, and for this population accurate gait assessment is central to preserving walking function, since gait deteriorates measurably by mid-adulthood in a quarter to half of adults with CP.
The Rodda and Graham classification system quantifies sagittal-plane gait deviations using ankle and knee z-scores derived from 3D Instrumented Gait Analysis (3D-IGA), but 3D-IGA is expensive and limited to large regional centers, while the widely available alternative, observational assessment, shows only moderate inter-rater agreement that drops further for less experienced clinicians.
We developed a markerless gait analysis pipeline that quantifies Rodda and Graham knee and ankle z-scores directly from single-view clinical gait videos.
Across 1{,}058 bilateral limb samples from 529 trials of 152 children (88 male, 63 female; age 12.1 $\pm$ 4.0 years; 60 distinct primary diagnoses, cerebral palsy the most common at $n = 54$), the sagittal-view model achieved $R^2 = 0.80 \pm 0.02$ and CCC $= 0.89 \pm 0.02$ for knee z-scores and $R^2 = 0.57 \pm 0.02$ and CCC $= 0.72 \pm 0.02$ for ankle z-scores against 3D-IGA.
Binary screening for excess knee flexion from predicted z-scores achieves AUROC $= 0.88$, correctly identifying 83\% of affected children, and applying Rodda and Graham classification rules yields $43 \pm 1\%$ 7-class accuracy with macro-AUROC $= 0.78 \pm 0.01$, ankle prediction error remaining the primary bottleneck.
Beyond cross-sectional screening, the continuous nature of predicted z-scores supports longitudinal trajectory tracking across clinical visits, providing a quantitative substrate for monitoring disease progression and treatment response that observational rating scales cannot offer.
These results demonstrate the feasibility of video-based knee z-score estimation, binary excess-flexion screening, and longitudinal trajectory tracking, offering a path toward scalable, objective gait assessment in low-resource clinical settings.
\end{abstract}


\section{Introduction}
\label{sec:intro}

Cerebral Palsy (CP) is a group of motor disorders caused by irreversible but non-progressive damage to the developing brain before, during, or shortly after birth~\citep{rosenbaum_report_2007}.
The condition is the most common cause of life-long physical disability in most developed countries, and affects the coordination and posture of approximately 2 per 1000 live births, with greater incidence in the developing world~\citep{oskoui_update_2013}.
Approximately 75\% of children with CP are ambulatory, encompassing children who walk independently through to those requiring assistive devices for household or community ambulation~\citep{armand_gait_2016}.
Walking ability directly predicts social participation and quality of life~\citep{vameghi_walking_2023}.
However, gait function is not stable as children age. 
Between a quarter and half of adults with CP experience measurable walking decline by mid-adulthood~\citep{graham_musculoskeletal_2021}, driven by progressive musculoskeletal deformity, spasticity, and soft tissue contracture that compound over the course of life~\citep{rosenbaum_report_2007}.
Therefore, for this large population, preserving and optimizing walking function is the central goal of treatment~\citep{keeratisiroj_prognostic_2018}.

\begin{figure*}[!t]
  \centering
  \includegraphics[width=\linewidth]{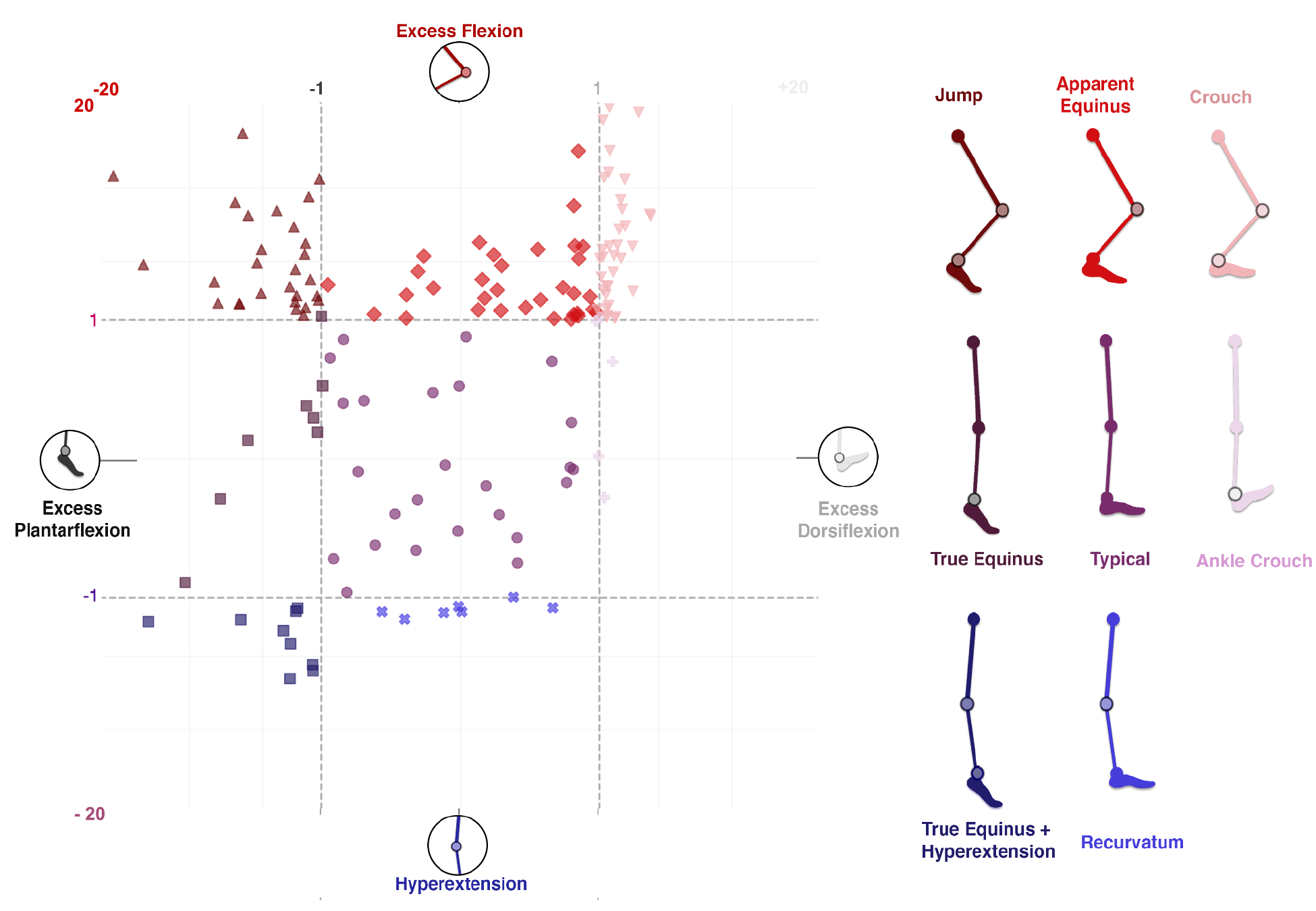}
  \caption{Rodda and Graham z-score space. The horizontal axis represents the ankle z-score (negative: excess plantarflexion; positive: excess dorsiflexion) and the vertical axis represents the knee z-score (negative: hyperextension; positive: excess flexion). The gray region represents z-scores between $-1$ and $+1$, considered within the normal range. Silhouettes depict each gait class.}
  \label{fig:rodda_graham_2d}
\end{figure*}

By analyzing a child's gait, clinicians make individualized treatment decisions, including prescribing therapy for muscle spasticity, surgically lengthening shortened muscles, and fitting orthoses to improve movement quality~\citep{armand_gait_2016, kay_effect_2000}.
The gold standard for gait analysis is 3D-Instrumented Gait Analysis (3D-IGA), which tracks anatomical landmarks in 3D space using multiple cameras and reflective markers to compute objective gait metrics, quantify gait deviations, and identify atypical gait patterns.~\citep{stebbins_clinical_2023, wren_efficacy_2011}.
A common system used to classify atypical gait patterns among individuals with CP is the Rodda and Graham classification, which categorizes children with CP by how far their knee and ankle angles deviate from typically developing controls~\citep{rodda_classification_2001, sangeux_sagittal_2015}.
The deviations span opposing directions, the knee may be hyperextended or excessively flexed, and the ankle may be excessively dorsiflexed or plantarflexed~\citep{rodda_sagittal_2004}.
Using kinematics derived from 3D-IGA, the 7 Rodda and Graham patterns, each with distinct clinical treatment pathways~\citep{rodda_classification_2001, rodda_sagittal_2004, kruger_shriners_2024} (\autoref{fig:rodda_graham_2d}) can be objectively identified ~\citep{rodda_classification_2001, rodda_sagittal_2004}.
However, 3D-IGA is expensive and geographically limited to specialized, often academic, regional centers that can afford the expensive hardware, and technical and clinical specialists to operate the systems and interpret the complex data they produce~\citep{stebbins_clinical_2023, armand_gait_2016}.
Rodda and Graham classification through observational assessment is low cost but shows only moderate-to-substantial agreement with kinematic ground truth ($\kappa = 0.67$ for experienced raters) that drops markedly for less experienced clinicians ($\kappa = 0.37$)~\citep{kim_reliability_2011, toro_review_2003, eastlack_interrater_1991}.
To address the gap between expensive gold-standard 3D-IGA and low-cost but unreliable observational assessment in children with CP, there is an urgent need for low-cost and objective gait quantification methods that can be deployed beyond the large regional centers where 3D-IGA is available.

Early work used deep learning to demonstrate the feasibility of automating extraction of clinically relevant metrics from data produced by multi camera 3D-IGA, multicamera pose estimation and monocular pose estimation systems.
Mazidi \textit{et al.}.~\citet{mazidi_application_2024} employed LSTM and attention-based networks to classify  Rodda and Graham patterns from 3D-IGA data in 317 CP patients, achieving 73\% classification accuracy across 4 Rodda and Graham classifications true equinus, apparent equinus, jump gait, and crouch gait, while Kim \textit{et al.}.~\citet{kim_deep-learning_2022} automated detection of gait cycle events, identifying initial-contact events with 89.7\% sensitivity ($\pm$16\,ms) from foot 3D-IGA data in 363 children with CP.
With the recent introduction of markerless pose estimation methods, researchers have demonstrated the feasibility of quantifying gait patterns from video data alone.
Kidzinski \textit{et al.}.~\citet{kidzinski_deep_2020} used the OpenPose pose estimator ~\citep{cao_openpose_2019} combined with deep learning to predict gait metrics including walking speed ($r = 0.73$), cadence ($r = 0.79$), and knee flexion angle at maximum extension ($r = 0.83$) from video, claiming to reach ``the theoretical limits imposed by natural within-subject variability''.
Pantzar-Castilla \textit{et al.}.~\citet{pantzar-castilla_feasibility_2024} used 2D markerless methods with RGBD cameras to quantitatively assess Rodda and Graham classification in 20 Swedish CP registry participants.
Zhao \textit{et al.}.~\citet{zhao_motor_2023} used Spatiotemporal Graph Convolutional Networks~\citep{yan_spatial_2018} to predict Gross Motor Function Classification System (GMFCS) levels in children with CP using 2D pose data captured from monocular video with 76.6\% accuracy.
Azhand \textit{et al.}.~\citet{azhand_algorithm_2021} quantified spatiotemporal gait parameters from monocular video that had a strong correlation (ICC $> 0.95$) with gait parameters acquired from clinical-grade pressure mats (e.g., GaitRite).
Together, these studies suggest that video-based gait analysis can approximate laboratory-grade assessments while substantially reducing infrastructure requirements.
Single-view video methods are especially relevant given the prevalence of smartphone cameras, which could enable automated gait screening without specialized hardware.

Although many studies have demonstrated the feasibility of video-based CP gait analysis, these techniques have not been validated for the full Rodda and Graham classification at scale, which has direct implications for the implementation of intervention strategies.
Of the single-camera video-based methods that exist, Zhao \textit{et al.}. \citet{zhao_motor_2023} predicted GMFCS levels in children with CP from monocular video to classify overall functional mobility severity from full independence (Level I) to full wheelchair dependence (Level V) ~\citep{palisano_development_1997}.
However, GMFCS level is not used to prescribe specific treatment decisions which require joint-level measurement from gait analysis~\citep{ounpuu_variation_2015}.
Pantzar \textit{et al.}.\ evaluate a small population, only 20 children with CP, they also utilize a RGBD (D for depth sensing) camera which is not readily available in most settings and ultimately do not automate score generation, instead depending on traditional evaluations of knee and ankle angles to identify classifications, which presents limitations on practical ability to identify gait classifications at scale, objectively.

This work evaluates Rodda and Graham ankle and knee z-score regression across eight camera viewpoints from clinical gait trial videos recorded during routine clinical visits, identifies sagittal view as the optimal recording angle, and demonstrates the feasibility of quantifying z-scores from this single view.
We further evaluate the clinical utility of predicted z-scores for clinical tasks like binary screening of excess knee flexion and for full 7-class Rodda and Graham gait classification, characterizing limitations in our work caused by regression error and mitigation strategies for improved performance.
The most consequential downstream use of such a system is longitudinal monitoring of gait impairments.
Continuous z-scores extracted from routine clinic video enable tracking of individual gait trajectories across visits to detect disease progression and treatment response at the patient level, a capability that the existing observational alternative fundamentally cannot provide.
These analyses represent a step toward developing accessible, scalable gait assessment systems for clinical decision support across the broader population of children with gait abnormalities seen in low-resource clinical settings, where Rodda and Graham deviations are diagnostically informative.

\section{Methods and Materials}
\label{sec:methods}

\subsection{Ethics Statement}
This study was approved by the Emory University Institutional Review Board (Protocol \#2024P007628) and the Shriners Children's Institutional Review Board (Protocol \#PHL2305).

\subsection{Dataset and Participants}
\label{sec:dataset}

\begin{figure*}[!t]
  \centering
  \includegraphics[width=\textwidth]{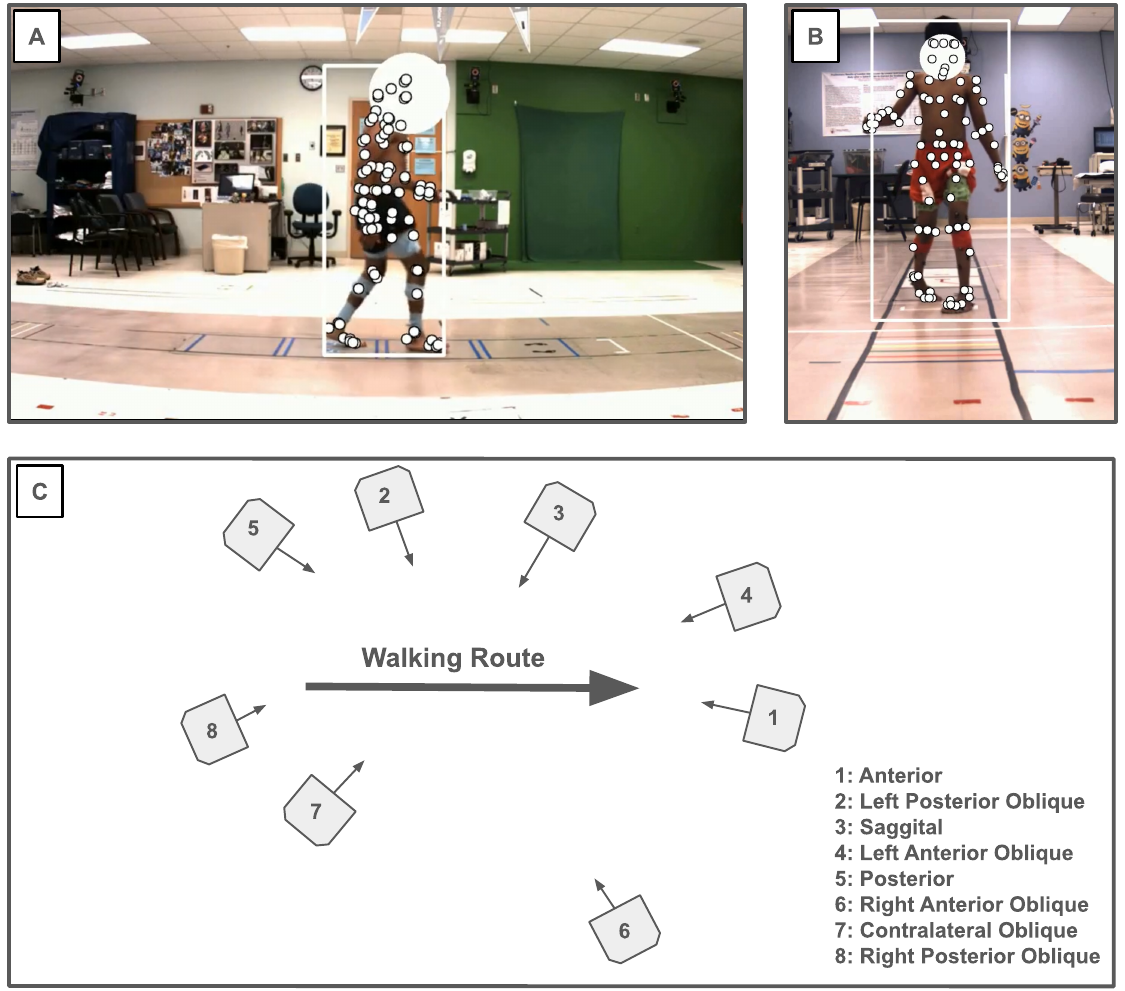}
  \vspace{0.5em}
  \includegraphics[width=\textwidth]{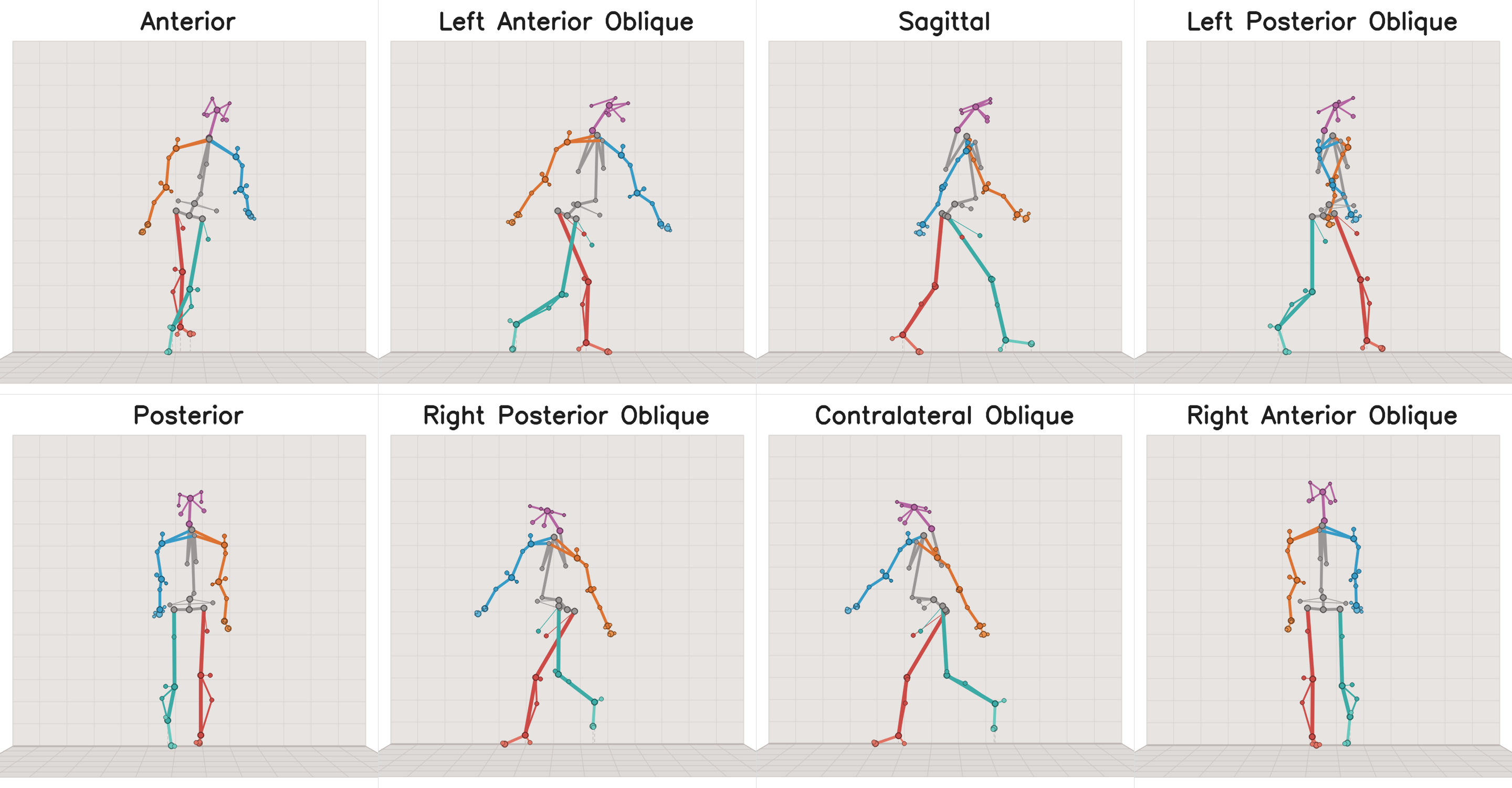}
  \caption{(a) The sagittal video stream. (b) The frontal video stream. (c) Experimental setup mockup showing the relative positioning of the multi-view recording array used during trials. (d) Monocular 3D pose estimation results across all eight camera viewpoints for a representative child.}
  \label{fig:pose_extraction}
\end{figure*}

Our video data were collected from Shriners Children's Philadelphia Motion Analysis Center during routine clinical visits and we had access to data from September 2023 to March-2026.
A convenience sample of 152 children (88 male, 63 female; age 12.1 $\pm$ 4.0 years, range 3--22; height 143.9 $\pm$ 20.6 cm; weight 46.6 $\pm$ 22.8 kg) representing 60 distinct primary diagnoses was evaluated using Rodda and Graham gait classification during routine 3D-IGA visits.
Cerebral palsy was the most common diagnosis (n = 54, 35.8\%), followed by arthrogryposis (n = 11, 7.3\%), talipes equinovarus (n = 10, 6.6\%), and spina bifida (n = 5, 3.3\%), with the remaining participants spanning a long tail of rare neurological, musculoskeletal, and genetic conditions (full breakdown in \nameref{S1_Table} of the Supplementary section).
All participants were ambulatory and underwent the same 3D-IGA protocol regardless of underlying diagnosis, yielding 529 gait trials with time-synchronized marker-based 3D kinematic data and video recordings from eight camera viewpoints (\autoref{fig:pose_extraction}, \autoref{tab:camera_views}).
Each trial captures a single walking bout from one side of the room to the other.
Trials per child range from 1 to 12 (mean 3.5).

\begin{table}[!t]
\centering
\caption{Camera viewpoints and trial counts. Left/Right refer to the patient's anatomical sides. $^\dagger$Fewer trials due to incomplete capture from this viewpoint.}
\label{tab:camera_views}
\small
\begin{tabular}{lrr}
\toprule
\textbf{View Name} & \textbf{No.\ Trials} & \textbf{No.\ Subjects} \\
\midrule
Anterior                & 529 & 152 \\
Left Anterior Oblique$^\dagger$   & 526 & 151 \\
Sagittal                & 529 & 152 \\
Left Posterior Oblique   & 529 & 152 \\
Posterior                & 529 & 152 \\
Right Posterior Oblique$^\dagger$  & 503 & 146 \\
Contralateral Oblique$^\dagger$    & 526 & 151 \\
Right Anterior Oblique   & 529 & 152 \\
\bottomrule
\end{tabular}
\end{table}

Rodda and Graham z-scores were computed from 3D-IGA for every trial using a clinician-selected gait cycle, regardless of the child's diagnosis, since the z-score is a kinematic deviation measure rather than a diagnosis-specific label.
A gait cycle is the sequence of motions that occurs from the initial contact of one heel to the next consecutive heel strike of the same foot, representing one full stride.
Rodda and Graham z-scores quantify how far a child's sagittal-plane knee and ankle angles deviate from typically developing controls.
For each joint $j \in \{\text{ankle},\, \text{knee}\}$, the z-score is computed over mid-stance (the 20--45 percentile window of the gait cycle, chosen to capture weight-bearing knee extension and ankle dorsiflexion while avoiding the transient loading response at initial contact~\citep{sangeux_sagittal_2015, kruger_shriners_2024}):
\begin{equation}
    z_j = \frac{\bar{\theta}_j^{[20\%, 45\%]} - \mu_j^{\mathrm{TD}}}{\sigma_j^{\mathrm{TD}}}
    \label{eq:zscore}
\end{equation}
where $\bar{\theta}_j^{[20\%, 45\%]}$ is the patient's mean sagittal joint angle over that window, and $\mu_j^{\mathrm{TD}}$ and $\sigma_j^{\mathrm{TD}}$ are the mean and standard deviation from a typically developing (TD) normative cohort.
Z-scores outside $[-1, +1]$ are considered non-normative, and thresholding along both joint axes defines seven Rodda and Graham gait classifications (\autoref{tab:rodda_regions}).
Z-scores are evaluated bilaterally where each trial yields separate ankle and knee z-scores for the left and right limbs, derived from the corresponding 3D-IGA.
To train a single model for both sides, we mirror the skeleton for one limb so that all samples share a canonical left-side orientation, yielding 1{,}058 limb-level samples from 529 trials per viewpoint.
Each z-score is computed from a single gait cycle selected by a clinician during the 3D-IGA session, typically when the child's foot contacts the force plate and reflects one representative cycle in the trial.
\autoref{fig:rodda_graham_2d} shows the overall distribution of ankle z-scores, ranging from $-19$ to $+5$, and knee z-scores, ranging from $-9$ to $+20$, spanning the full spectrum of Rodda and Graham gait patterns~\citep{rodda_classification_2001}.

\begin{table}[!t]
\centering
\caption{Rodda and Graham gait class regions defined by knee ($z_k$) and ankle ($z_a$) z-score thresholds at $\pm 1$. See Equation~\ref{eq:zscore} for the z-score definition.}
\label{tab:rodda_regions}
\small
\begin{tabular}{rlll}
\toprule
& \textbf{Class} & \textbf{Knee} & \textbf{Ankle} \\
\midrule
0. & Normal & $|z_k| < 1$ & $|z_a| < 1$ \\
1. & True Equinus & $z_k < 1$ & $z_a < -1$ \\
2. & Jump Gait & $z_k > 1$ & $z_a < -1$ \\
3. & Apparent Equinus & $z_k > 1$ & $|z_a| < 1$ \\
4. & Crouch & $z_k > 1$ & $z_a > 1$ \\
5. & Ankle Crouch & $|z_k| < 1$ & $z_a > 1$ \\
6. & Recurvatum & $z_k < -1$ & $|z_a| < 1$ \\
\bottomrule
\end{tabular}
\end{table}

\subsection{Video-Based Gait Analysis Pipeline}

\begin{figure*}[p]
  \begin{adjustwidth}{-1.75in}{0in}
  \centering
  \includegraphics[width=7.5in,height=8.75in,keepaspectratio]{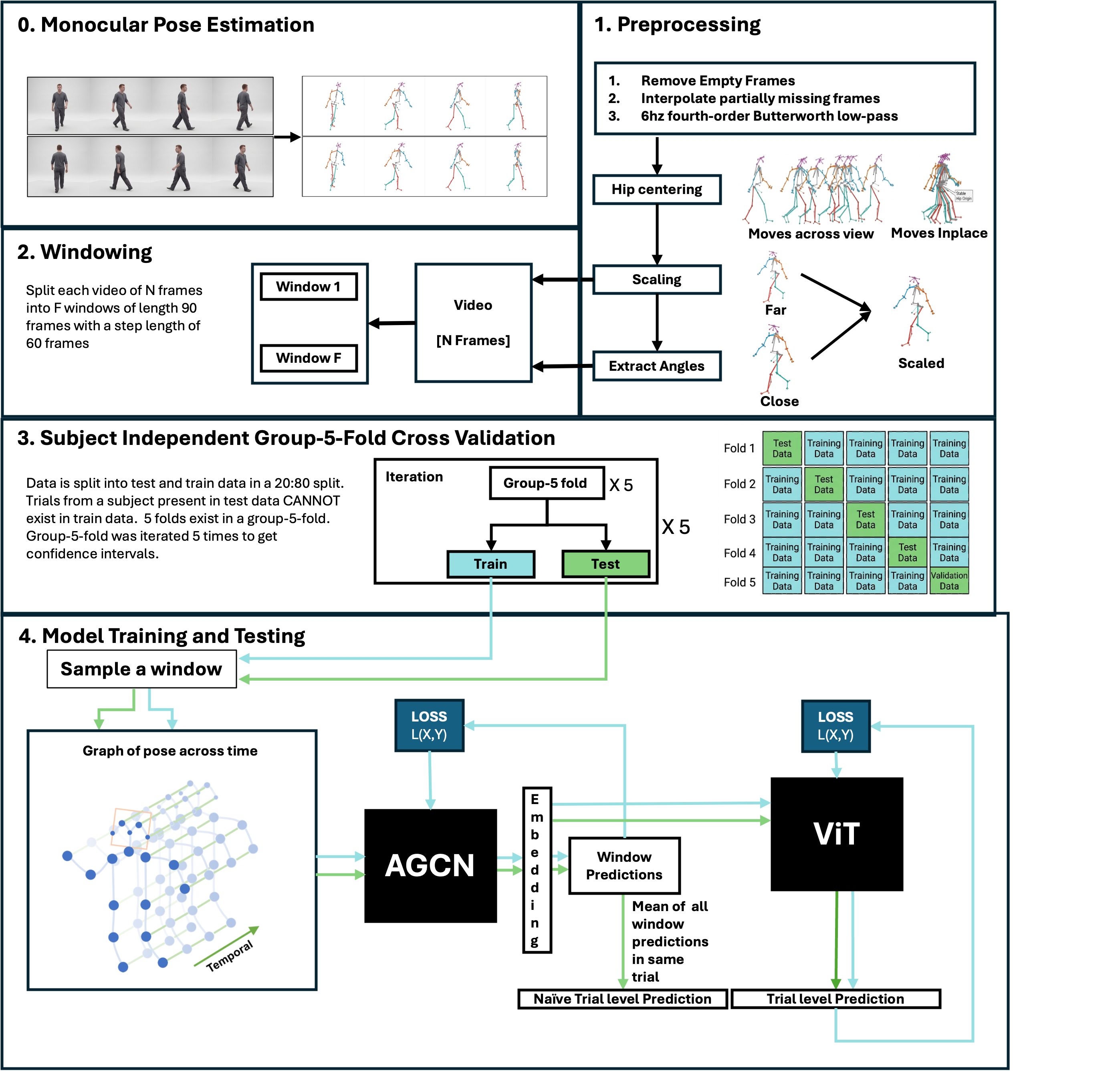}
  \caption{Overview of the experimental pipeline. Monocular 3D pose estimation extracts keypoints from clinical gait video, which are cleaned and represented as raw coordinates or derived joint angles. Both representations are windowed and processed by deep learning models (DCL, ST-GCN, AGCN, and AGCN+ViT) under participant-wise 5-fold cross-validation for ankle and knee z-score regression.}
  \label{fig:pipeline}
  \end{adjustwidth}
\end{figure*}

We developed a pipeline that regresses knee and ankle z-scores, derived from 3D instrumented gait analysis, directly from the monocular clinical gait video (\autoref{fig:pipeline}).
To determine which camera viewpoint yielded the strongest predictive performance, we applied the pipeline independently to each of the eight viewpoints in the recording array.
Monocular 3D pose estimation first extracts raw keypoints from each video frame (\autoref{fig:pipeline} 0. Monocular Pose estimation), which are cleaned, low-pass filtered, and scaled to normalize for stature(\autoref{fig:pipeline} 1. Preprocessing).
The cleaned keypoints are evaluated in two representations: raw spatial coordinates and derived joint angle time series(\autoref{fig:pipeline} 1. Preprocessing).
As a natural first step, we evaluate a non-learned biomechanical baseline that computes z-scores directly from the monocular-derived angles (as detailed in \autoref{eq:zscore} ) to establish the performance floor any learned model must improve upon.
For deep learning analysis both representations are segmented into 1.5 second sliding windows, each inheriting the trial-level z-score from 3D-IGA as its regression target (\autoref{fig:pipeline} 2. Windowing).
Three deep learning models (DCL~\citep{ordonez_deep_2016}, ST-GCN~\citep{yan_spatial_2018}, and AGCN~\citep{shi_two_2019}) are trained on window-level features using participant-wise 5-fold cross-validation(\autoref{fig:pipeline} 3. Subject Independent Group-5-Fold Cross Validation).
For each trial window, the base models (DCL, ST-GCN, and AGCN) produce a window-level z-score prediction(\autoref{fig:pipeline} 4. Model Training and Testing).
A naive trial-level z-score estimate is obtained by averaging these window-level predictions across all windows in the trial (\autoref{fig:pipeline} 4. Model Training and Testing).
In addition, a fourth vision transformer (ViT) model~\citep{dosovitskiy_image_2021} is trained to produce trial-level scores by treating the window embeddings extracted from the
penultimate layer of the best-performing base model as tokens (\autoref{fig:pipeline} 4. Model Training and Testing).
Our AGCN+ViT design hierarchically and adaptively aggregates information across the full sequence of window embeddings, allowing it to model long-range temporal dependencies and output a single trial-level z-score (\autoref{fig:pipeline} 4. Model Training and Testing).
Each component is described in detail in the subsections below.

\subsubsection{Pose Estimation and Preprocessing}
\label{sec:preprocessing}

We estimated full-body 3D joint keypoints from each video frame using MeTRAbs~\citep{sarandi_metrabs_2021, sarandi_dozens_2023}, an off-the-shelf monocular 3D pose estimator that predicts joint positions in real-world metric coordinates (meters) rather than pixel coordinates.
The 2023 version~\citep{sarandi_dozens_2023} is trained jointly on dozens of motion capture datasets, each using a different skeleton definition, and learns a shared internal representation of human pose that can be decoded into any of 23 output skeleton formats.
We use the BML-MoVi output format~\citep{ghorbani_movi_2021}, yielding 87 keypoints across the full body.
We apply a multi-stage cleaning and normalization pipeline to the raw keypoint sequences (See \autoref{fig:pipeline} 1. Preprocessing).
First, we remove frames with all-zero keypoint coordinates (failed detections due to individual not being within view).
Second, we linearly interpolate partially  missing keypoints due to partial occlusion resulting in failed detection of body parts.
Third, We apply a fourth-order Butterworth low-pass filter with a 6\,Hz cutoff independently to each coordinate time series to attenuate high-frequency pose-estimation jitter while preserving gait dynamics below 3--4\,Hz~\citep{antonsson_frequency_1985}.
We hip-center all keypoint coordinates by subtracting the mid-hip position from every keypoint at each frame, placing the pelvis origin at $(0,0)$ and removing global translation from the child's position in the camera field of view to standardize the coordinate system of data for model training.
To account for variation in participant stature and camera distance, we rescale coordinates so that the smoothed backneck-to-sternum distance matches a fixed reference length, using a 30-frame moving average of the reference segment to reduce frame-to-frame scaling jitter caused by pose estimation error in ratios of body segments in relation to each other.

After cleaning and normalization, we evaluate two input representations derived from the preprocessed keypoints.
The first representation uses the preprocessed $(x,y,z)$ keypoint coordinates, as described above, directly as spatial features.
The second representation derives $F{=}24$ joint angle time series from the cleaned keypoints by deriving 3D rotation matrices for selected joints and their corresponding segments, where segments are a line described by the connection between two keypoints selected by us from among the 87 keypoint set, following the conventions established by the International Society of Biomechanics~\citep{wu_isb_2002}.
From the 87-keypoint set, we select 17 anatomical landmarks: bilateral hip, knee (lateral and medial), ankle (lateral and medial), heel, and toe joint centers, plus mid-hip, backneck (C7), and sternum.
For each body segment $s$, we construct a local right-handed coordinate frame $[\mathbf{e}_x^s,\, \mathbf{e}_y^s,\, \mathbf{e}_z^s]$ from the surrounding keypoints, where $\mathbf{e}_x$ is the medio-lateral axis, $\mathbf{e}_y$ is the longitudinal (proximal-distal) axis, and $\mathbf{e}_z$ is the anterior-posterior axis (computed as $\mathbf{e}_z = \mathbf{e}_x \times \mathbf{e}_y$).
The four segment frames are defined as follows:
\begin{itemize}
\item \textbf{Pelvis:} $\mathbf{e}_x = \overrightarrow{\text{lhip} \to \text{rhip}}$ (medio-lateral), $\mathbf{e}_y = \overrightarrow{\text{mhip} \to \text{backneck}}$ (longitudinal).
\item \textbf{Thigh:} $\mathbf{e}_y = \overrightarrow{\text{knee} \to \text{hip}}$ (longitudinal), $\mathbf{e}_x = \overrightarrow{\text{knee}_\text{med} \to \text{knee}_\text{lat}}$ (medio-lateral).
\item \textbf{Shank:} $\mathbf{e}_y = \overrightarrow{\text{ankle} \to \text{knee}}$ (longitudinal), $\mathbf{e}_x = \overrightarrow{\text{ankle}_\text{med} \to \text{ankle}_\text{lat}}$ (medio-lateral).
\item \textbf{Foot:} $\mathbf{e}_y = \overrightarrow{\text{heel} \to \text{toe}}$ (progression), $\mathbf{e}_x = \overrightarrow{\text{ankle}_\text{med} \to \text{ankle}_\text{lat}}$ (medio-lateral).
\end{itemize}
Each axis is normalized to unit length, and $\mathbf{e}_z$ is obtained via cross product to complete the orthonormal basis.
The rotation matrix for each segment is then $\mathbf{R}_s = [\mathbf{e}_x^s \;\; \mathbf{e}_y^s \;\; \mathbf{e}_z^s]$.

Joint angles are computed from the relative rotation between adjacent segment frames.
For a proximal segment with rotation $\mathbf{R}_p$ and a distal segment with rotation $\mathbf{R}_d$, the relative rotation matrix is:
\begin{equation}
\mathbf{R}_{\text{rel}} = \mathbf{R}_p^\top \, \mathbf{R}_d
\label{eq:rel_rotation}
\end{equation}
This relative rotation is decomposed into XYZ Euler angles $(\alpha, \beta, \gamma)$ in degrees, where $\alpha$ corresponds to flexion/extension, $\beta$ to adduction/abduction, and $\gamma$ to internal/external rotation.
This procedure is applied at four joints per limb (hip, knee, ankle, and a knee-medial-to-mid-hip reference), bilaterally, yielding $8 \times 3 = 24$ angle channels per frame (\autoref{tab:angle_channels}).
Joint angles are the standardized biomechanical representation for reporting human joint motion~\citep{wu_isb_2002} and are inherently invariant to camera distance and participant stature, unlike raw keypoint coordinates which encode absolute spatial positions that vary with both.
In clinical gait analysis, joint angles are the primary representation format for movement data~\citep{perry_gait_2010, baker_gait_2006} where marker or keypoint positions are collected as an intermediate step from which joint angles are computed via segment-fixed coordinate systems and Euler decomposition~\citep{davis_gait_1991, kadaba_measurement_1990}.
The Rodda and Graham z-scores that serve as our regression targets are themselves derived from sagittal-plane joint angles~\citep{sangeux_sagittal_2015}, making joint angles the natural input feature for a learned regressor.
Both spatial and joint angle representations are segmented into fixed-length analysis windows with a sliding window of 1.5 seconds and a stride of 1 second (33\% overlap) (\autoref{fig:pipeline} 2. Windowing).
Each 90-frame window captures approximately 1--2 complete gait cycles and 1.5 seconds at typical cadence in children with CP.
We discard windows shorter than 1.5 seconds at the end of a trial.
Each window inherits trial-level ankle and knee z-scores from 3D-IGA as window-level regression targets (or labels).

\begin{table}[!t]
\centering
\caption{The 24 joint angle channels derived from segment-to-segment rotations. Each joint produces three rotation components. L/R = left/right.}
\label{tab:angle_channels}
\footnotesize
\setlength{\tabcolsep}{4pt}
\begin{tabular}{@{}l l p{0.58\linewidth}@{}}
\toprule
\textbf{Joint} & \textbf{Angles} & \textbf{Anatomical Reference Landmarks} \\
\midrule
R Hip       & flex, add, rot       & R hip joint center, mid-pelvis, R knee joint center \\
R Knee      & flex, add, rot       & R knee joint center, R hip joint center, R lateral malleolus, R medial femoral epicondyle \\
R Ankle     & dorsiflex, inv, abd  & R lateral malleolus, R knee joint center, R medial malleolus, R forefoot, R calcaneus \\
R Knee-MHip & flex, add, rot       & R medial femoral epicondyle, mid-pelvis, R hip joint center, L hip joint center \\
L Hip       & flex, add, rot       & L hip joint center, mid-pelvis, L knee joint center \\
L Knee      & flex, add, rot       & L knee joint center, L hip joint center, L lateral malleolus, L medial femoral epicondyle \\
L Ankle     & dorsiflex, inv, abd  & L lateral malleolus, L knee joint center, L medial malleolus, L forefoot, L calcaneus \\
L Knee-MHip & flex, add, rot       & L medial femoral epicondyle, mid-pelvis, L hip joint center, R hip joint center \\
\bottomrule
\end{tabular}
\end{table}

\subsubsection{Gait Feature Extraction with AGCN}
\label{sec:models}

We use Adaptive Graph Convolutional Networks (AGCN)~\citep{shi_two_2019} as the feature extractor for trial-level z-score regression using ViT.
AGCN has been shown to be effective for gait analysis in neurological conditions such as Parkinson's disease~\citep{kwon_explainable_2023}.
AGCN~\citep{shi_two_2019} represents the body as a spatiotemporal graph, and is an improvement of the Spatiotemporal Graph Convolutional Network (ST-GCN)~\citep{yan_spatial_2018} backbone, which uses Graph Convolutional Networks (GCN)~\citep{kipf_semi_2017} to model joint relationships in both space and time.
This graph is constructed by treating each joint as a node, and the anatomical links between joints, such as hip-to-knee or knee-to-ankle, define the initial connections between nodes.
When using joint angle features, the graph has $V{=}24$ nodes (one per angle channel) and when using raw keypoint coordinates, the graph has $V{=}87$ nodes (one per keypoint).
Edges follow the skeletal kinematic chain (hip-knee-ankle per limb), with additional connections between bilateral hip centers via the pelvis.
This graph is defined fully using a matrix $\mathbf{A} \in \{0,1\}^{V \times V}$ that we call an adjacency matrix.
The network comprises four stacked blocks, each combining a Temporal Convolutional Network (TCN) with a GCN.
Hence, each block applies spatial graph convolution followed by temporal convolution.
In ST-GCN~\citep{yan_spatial_2018}, the adjacency matrix is fixed, meaning the same joint relationships are assumed for every child.
However, pediatric gait deviations are heterogeneous and the same Rodda and Graham classification can arise from very different visual movement patterns, both within CP~\citep{pandey_crouch_2023} and across other diagnoses sharing similar kinematic deviations.
The key benefit of using AGCN~\citep{shi_two_2019} is that it self-generates the adjacency matrix weights by projecting the input window through learnt convolutional layers.
These joint-level relationships are learned together with the temporal information aggregated with the TCNs.
As a result, when the model analyzes a gait trial, it can place greater emphasis on joint relationships that are most informative for that individual and temporal region while reducing the influence of less informative relationships.
Thus, AGCNs learn individual-specific graph topologies on a per-window basis, allowing the model to discover which joint relationships are most informative for each trial rather than relying on a single fixed skeletal structure.
Please refer to the original paper~\citep{shi_two_2019} for mathematical details.
Each model is trained independently on both input representations (raw keypoint coordinates and derived joint angles) to evaluate whether biomechanical feature engineering improves prediction.

\subsubsection{Trial-Level Aggregation}
\label{sec:aggregation}

Window-level models produce one z-score prediction per 90-frame window. Aggregating these into a single trial-level estimate requires a pooling strategy.
We compare two approaches: a naive average pooling baseline and a hierarchical attention-based aggregation using a ViT~\citep{dosovitskiy_image_2021}.

\paragraph{Average Pooling}
The simplest aggregation assumes that all windows are equally informative. Trial-level z-scores for the knee and ankle are computed by taking the arithmetic mean of their respective window-level predictions:

\[
\hat{z}_{j,\text{trial}} = \frac{1}{N} \sum_{i=1}^{N} \hat{z}_{j,i},
\qquad j \in \{\text{knee}, \text{ankle}\}
\]

where \(\hat{z}_{j,i}\) is the predicted z-score for the \(i\)-th window for joint \(j\), and \(N\) is the number of windows in the trial.

\paragraph{Vision Transformer Aggregation}

As described in Section \nameref{sec:dataset}, the ground-truth z-scores derive from a single clinician-selected gait cycle (\autoref{eq:zscore}).
Training models using window level samples creates a weak supervision where each sliding window belonging to a trial, inherits the trial label despite cycle-to-cycle kinematic variability which can be as much as 2\textdegree at the knee~\citep{tabard-fougere_clinical_2022}.
We therefore attempt to model the long-range dependencies across the entire trial with a self-attention mechanism at once by adopting a two-stage training procedure inspired by the  ViT~\citep{dosovitskiy_image_2021} (\autoref{fig:pipeline} 4.Model Training and Testing).
The trained AGCN~\citep{shi_two_2019} backbone is frozen and its penultimate-layer embeddings are extracted for each window, becoming individual tokens in the sequence of all windows in the trial.
A learnable CLS token is prepended to the sequence, and the combined sequence is passed through the ViT.
The CLS token output is projected by a linear layer to produce the trial-level z-score.

\subsection{Experimental Protocol}

\subsubsection{Baseline Models}

To evaluate whether the adaptive graph topology of AGCN~\citep{shi_two_2019} is necessary, we compare against two baseline architectures that make progressively simpler structural assumptions.
ST-GCN~\citep{yan_spatial_2018} shares the same four-block TCN+GCN architecture as AGCN~\citep{shi_two_2019} but uses a fixed skeletal adjacency matrix rather than attention-based adaptable graphical structure, testing whether a predefined graph, strictly following the joint and limb interactions based on human skeletal structure, suffices for z-score regression.
DCL~\citep{ordonez_deep_2016} does away with the graph structure entirely and treats the input as a single-channel pseudo image of shape $(T{=}90, F)$ and processes it through four stacked 2D convolutional layers (64 filters each, kernel size $(5, 1)$) followed by a two-layer LSTM~\citep{hochreiter_long_1997}, testing whether temporal dynamics alone carry sufficient predictive signal without explicit joint relationship modeling.
DCL is a widely adopted baseline for deep learning-based human activity recognition from sensor and motion-capture streams~\citep{ordonez_deep_2016, hammerla_deep_2016}, including kinematic gait analysis in neurological conditions~\citep{kwon_explainable_2023}, which sets this model as a baseline model for this work.
We additionally include a biomechanical approach-based baseline that bypasses representation learning entirely for each viewpoint. We compute Rodda and Graham knee and ankle z-scores directly from the cleaned monocular-derived joint angles using \autoref{eq:zscore} and average per-cycle z-scores within each trial to obtain a trial-level prediction.
This baseline tests whether the kinematic content of single-view markerless pose-estimated angles is, on its own, sufficient to recover the 3D-IGA-derived z-scores, isolating the contribution of deep learning models.

\subsubsection{Training}

We trained models using five repeated runs of Grouped 5-fold cross-validation.
Patient identifiers were used as grouping variables to ensure that all samples from a given child, including left and right limbs across all trials, were assigned to only one partition within each fold. 
The training, validation, and test sets were divided in a 3:1:1 ratio.
Hyperparameters were selected using Optuna-based optimization~\citep{akiba_optuna_2019} within a nested participant stratified group 5-fold. 
For AGCN~\citep{shi_two_2019} and ST GCN~\citep{yan_spatial_2018}, the search space included learning rate from $10^{-5}$ to $10^{-2}$ on a log scale and weight decay from $10^{-6}$ to $10^{-2}$ on a log scale. 
The configuration range also allowed graph convolution filters from 32 to 128, dense units from 256 to 1024, and dropout from 0.2 to 0.7. For ST GCN, the graph layer range was 2 to 6. 
For DCL~\citep{ordonez_deep_2016}, the search space included learning rate from $10^{-5}$ to $10^{-2}$ on a log scale, weight decay from $10^{-6}$ to $10^{-2}$ on a log scale, 32 to 128 convolutional filters, 64 to 256 LSTM units, 1 to 4 LSTM layers, 2 to 6 convolutional layers, filter sizes from 3 to 9, and dropout from 0.2 to 0.7.
The selected hyperparameters were then fixed and used for the full outer fold evaluation. All backbone models were trained with mean squared error loss, optimized with Adam~\citep{kingma_adam_2015}, for 100 epochs with batch size 16

\subsubsection{Evaluation}

For the best performing model type and views, we present results for the coefficient of determination ($R^2$), root mean squared error (RMSE), mean absolute error (MAE), concordance correlation coefficient (CCC), and systematic bias from Bland-Altman analysis.
CCC measures agreement between predicted and true values accounting for both correlation and systematic bias.
All metrics are reported as mean $\pm$ 95\% confidence interval across the 5 iterations.
To assess the clinical utility of continuous z-score predictions, we evaluate the models on two downstream classification tasks from the regression output.
First, we threshold predicted knee z-scores at $+1$ to obtain a binary classifier for excess knee flexion, evaluating with Area Under the Receiver Operating Curve (AUROC), Area Under the Precision Recall Curve (AUPRC), accuracy, F1, precision, recall, and specificity.
Second, we apply the Rodda and Graham classification rules (\autoref{tab:rodda_regions}) to the predicted ankle and knee z-scores jointly to classify each trial into one of seven gait patterns.
We further stratify regression performance by proximity to the $\pm 1$ classification boundary (easy vs.\ hard trials) to characterize how regression error translates into classification reliability.

\section{Results}
\label{sec:results}

\subsection{Multi-View Comparison}
\label{sec:multiview}

\autoref{tab:multiview_angles} reports $R^2$ across all eight camera viewpoints for (a)joint angle and keypoint features, respectively, at both trial (T) level using naive pooling and window (W) levels.
The sagittal view consistently yielded the highest $R^2$ across all models for both ankle and knee z-scores.
AGCN~\citep{shi_two_2019} achieves the best performance at every viewpoint in both window level and naively aggregated trial level, followed by DCL~\citep{ordonez_deep_2016} and ST-GCN~\citep{yan_spatial_2018}.
For trial-level knee z-scores under AGCN with average pooling, the contralateral oblique ($R^2 = 0.74$) and left anterior oblique ($R^2 = 0.72$) views approach sagittal performance ($R^2 = 0.77$).
Anterior and posterior views yield the lowest $R^2$ for both targets.
Across all viewpoints, knee z-scores are predicted more accurately than ankle z-scores.
Joint angle features outperform raw keypoint features across all viewpoints and models.
The Biomech baseline, which computes Rodda and Graham z-scores directly from single-view 2D joint angles without learning, yielded negative $R^2$ at every viewpoint (best: sagittal knee $R^2 = -1.30$, ankle $R^2 = -0.38$; worst: posterior knee $R^2 = -4.51$).
Based on these results, all subsequent analyses (detailed regression metrics, error stratification, classification, and calibration) use the sagittal view with joint angle features exclusively.

\begin{table*}[!t]
\centering
\caption{$R^2$ (mean $\pm$ SD) across camera viewpoints with average pooling. T = trial-level, W = window-level. Best trial-level result per viewpoint in \textbf{bold}. Views ordered by AGCN knee $R^2$. (a) Joint angle features, including a Biomech baseline that computes Rodda and Graham z-scores directly from single-view 2D joint angles. (b) Raw keypoint features.}
\label{tab:multiview_angles}
\label{tab:multiview_keypoints}
\footnotesize
\setlength{\tabcolsep}{3pt}
\textbf{(a) Joint Angle Features}\\[0.3em]
\begin{tabular*}{\textwidth}{@{\extracolsep{\fill}} ll cc cc cc cc}
\toprule
& & \multicolumn{2}{c}{Biomech} & \multicolumn{2}{c}{DCL} & \multicolumn{2}{c}{ST-GCN} & \multicolumn{2}{c}{AGCN} \\
\cmidrule(lr){3-4} \cmidrule(lr){5-6} \cmidrule(lr){7-8} \cmidrule(lr){9-10}
View & & Ankle & Knee & Ankle & Knee & Ankle & Knee & Ankle & Knee \\
\midrule
Sagittal         & T & $-.38$ & $-1.30$ & $.51{\scriptstyle\pm.04}$ & $.68{\scriptstyle\pm.03}$ & $.24{\scriptstyle\pm.07}$ & $.36{\scriptstyle\pm.04}$ & $\mathbf{.58}{\scriptstyle\pm.03}$ & $\mathbf{.77}{\scriptstyle\pm.01}$ \\
                 & W & NA & NA & $.43{\scriptstyle\pm.06}$ & $.70{\scriptstyle\pm.03}$ & $.14{\scriptstyle\pm.09}$ & $.45{\scriptstyle\pm.03}$ & $.52{\scriptstyle\pm.03}$ & $.80{\scriptstyle\pm.01}$ \\
\addlinespace
Contra.\ Oblique & T & $-.42$ & $-1.72$ & $.48{\scriptstyle\pm.06}$ & $.68{\scriptstyle\pm.02}$ & $.21{\scriptstyle\pm.08}$ & $.21{\scriptstyle\pm.08}$ & $\mathbf{.49}{\scriptstyle\pm.02}$ & $\mathbf{.74}{\scriptstyle\pm.01}$ \\
                 & W & NA & NA & $.38{\scriptstyle\pm.05}$ & $.63{\scriptstyle\pm.03}$ & $.15{\scriptstyle\pm.07}$ & $.25{\scriptstyle\pm.07}$ & $.42{\scriptstyle\pm.02}$ & $.71{\scriptstyle\pm.01}$ \\
\addlinespace
L Ant.\ Oblique  & T & $-.45$ & $-2.56$ & $.44{\scriptstyle\pm.07}$ & $.68{\scriptstyle\pm.03}$ & $-.02{\scriptstyle\pm.02}$ & $.41{\scriptstyle\pm.08}$ & $\mathbf{.44}{\scriptstyle\pm.02}$ & $\mathbf{.72}{\scriptstyle\pm.02}$ \\
                 & W & NA & NA & $.30{\scriptstyle\pm.06}$ & $.66{\scriptstyle\pm.02}$ & $-.04{\scriptstyle\pm.02}$ & $.40{\scriptstyle\pm.09}$ & $.34{\scriptstyle\pm.02}$ & $.72{\scriptstyle\pm.02}$ \\
\addlinespace
R Post.\ Oblique & T & $-.62$ & $-2.74$ & $.24{\scriptstyle\pm.03}$ & $.62{\scriptstyle\pm.02}$ & $.05{\scriptstyle\pm.05}$ & $.41{\scriptstyle\pm.09}$ & $\mathbf{.35}{\scriptstyle\pm.01}$ & $\mathbf{.65}{\scriptstyle\pm.01}$ \\
                 & W & NA & NA & $.12{\scriptstyle\pm.03}$ & $.52{\scriptstyle\pm.02}$ & $.04{\scriptstyle\pm.04}$ & $.42{\scriptstyle\pm.09}$ & $.32{\scriptstyle\pm.02}$ & $.59{\scriptstyle\pm.01}$ \\
\addlinespace
L Post.\ Oblique & T & $-.62$ & $-3.67$ & $.37{\scriptstyle\pm.04}$ & $.61{\scriptstyle\pm.02}$ & $.16{\scriptstyle\pm.05}$ & $.15{\scriptstyle\pm.05}$ & $\mathbf{.46}{\scriptstyle\pm.01}$ & $\mathbf{.63}{\scriptstyle\pm.02}$ \\
                 & W & NA & NA & $.26{\scriptstyle\pm.06}$ & $.60{\scriptstyle\pm.01}$ & $.13{\scriptstyle\pm.05}$ & $.19{\scriptstyle\pm.05}$ & $.39{\scriptstyle\pm.02}$ & $.65{\scriptstyle\pm.02}$ \\
\addlinespace
Anterior         & T & $-.62$ & $-4.04$ & $.26{\scriptstyle\pm.13}$ & $.48{\scriptstyle\pm.03}$ & $.04{\scriptstyle\pm.03}$ & $.27{\scriptstyle\pm.05}$ & $\mathbf{.38}{\scriptstyle\pm.03}$ & $\mathbf{.57}{\scriptstyle\pm.02}$ \\
                 & W & NA & NA & $.15{\scriptstyle\pm.14}$ & $.41{\scriptstyle\pm.03}$ & $.03{\scriptstyle\pm.02}$ & $.30{\scriptstyle\pm.04}$ & $.33{\scriptstyle\pm.03}$ & $.56{\scriptstyle\pm.02}$ \\
\addlinespace
R Ant.\ Oblique  & T & $-.53$ & $-4.29$ & $.17{\scriptstyle\pm.13}$ & $.47{\scriptstyle\pm.05}$ & $-.02{\scriptstyle\pm.04}$ & $.38{\scriptstyle\pm.03}$ & $\mathbf{.26}{\scriptstyle\pm.04}$ & $\mathbf{.54}{\scriptstyle\pm.01}$ \\
                 & W & NA & NA & $.09{\scriptstyle\pm.10}$ & $.46{\scriptstyle\pm.03}$ & $-.02{\scriptstyle\pm.04}$ & $.42{\scriptstyle\pm.03}$ & $.22{\scriptstyle\pm.04}$ & $.54{\scriptstyle\pm.01}$ \\
\addlinespace
Posterior        & T & $-1.11$ & $-4.51$ & $.28{\scriptstyle\pm.07}$ & $.43{\scriptstyle\pm.05}$ & $.06{\scriptstyle\pm.08}$ & $.37{\scriptstyle\pm.03}$ & $\mathbf{.35}{\scriptstyle\pm.01}$ & $\mathbf{.51}{\scriptstyle\pm.03}$ \\
                 & W & NA & NA & $.24{\scriptstyle\pm.07}$ & $.37{\scriptstyle\pm.07}$ & $.06{\scriptstyle\pm.09}$ & $.38{\scriptstyle\pm.04}$ & $.35{\scriptstyle\pm.01}$ & $.47{\scriptstyle\pm.02}$ \\
\bottomrule
\end{tabular*}
\\[1em]
\textbf{(b) Raw Keypoint Features}\\[0.3em]
\begin{tabular*}{\textwidth}{@{\extracolsep{\fill}} ll cc cc cc}
\toprule
& & \multicolumn{2}{c}{DCL} & \multicolumn{2}{c}{ST-GCN} & \multicolumn{2}{c}{AGCN} \\
\cmidrule(lr){3-4} \cmidrule(lr){5-6} \cmidrule(lr){7-8}
View & & Ankle & Knee & Ankle & Knee & Ankle & Knee \\
\midrule
Sagittal         & T & $-.10{\scriptstyle\pm.01}$ & $.31{\scriptstyle\pm.01}$ & $.03{\scriptstyle\pm.05}$ & $.06{\scriptstyle\pm.02}$ & $\mathbf{.39}{\scriptstyle\pm.05}$ & $\mathbf{.46}{\scriptstyle\pm.04}$ \\
                 & W & $-.13{\scriptstyle\pm.02}$ & $.38{\scriptstyle\pm.04}$ & $-.01{\scriptstyle\pm.04}$ & $.19{\scriptstyle\pm.04}$ & $.28{\scriptstyle\pm.06}$ & $.51{\scriptstyle\pm.04}$ \\
\addlinespace
Contra.\ Oblique & T & $-.01{\scriptstyle\pm.01}$ & $.24{\scriptstyle\pm.07}$ & $.09{\scriptstyle\pm.05}$ & $.13{\scriptstyle\pm.05}$ & $\mathbf{.37}{\scriptstyle\pm.06}$ & $\mathbf{.55}{\scriptstyle\pm.04}$ \\
                 & W & $-.02{\scriptstyle\pm.01}$ & $.24{\scriptstyle\pm.05}$ & $.05{\scriptstyle\pm.09}$ & $.20{\scriptstyle\pm.03}$ & $.31{\scriptstyle\pm.05}$ & $.55{\scriptstyle\pm.03}$ \\
\addlinespace
L Ant.\ Oblique  & T & $-.06{\scriptstyle\pm.02}$ & $.03{\scriptstyle\pm.05}$ & $-.07{\scriptstyle\pm.03}$ & $.12{\scriptstyle\pm.04}$ & $\mathbf{.09}{\scriptstyle\pm.03}$ & $\mathbf{.52}{\scriptstyle\pm.06}$ \\
                 & W & $-.07{\scriptstyle\pm.02}$ & $.03{\scriptstyle\pm.05}$ & $-.11{\scriptstyle\pm.04}$ & $.19{\scriptstyle\pm.04}$ & $.02{\scriptstyle\pm.04}$ & $.53{\scriptstyle\pm.06}$ \\
\addlinespace
R Post.\ Oblique & T & $-.10{\scriptstyle\pm.06}$ & $.07{\scriptstyle\pm.05}$ & $-.08{\scriptstyle\pm.01}$ & $.15{\scriptstyle\pm.03}$ & $-.04{\scriptstyle\pm.04}$ & $.48{\scriptstyle\pm.03}$ \\
                 & W & $-.12{\scriptstyle\pm.09}$ & $.05{\scriptstyle\pm.06}$ & $-.09{\scriptstyle\pm.01}$ & $.15{\scriptstyle\pm.04}$ & $-.10{\scriptstyle\pm.05}$ & $.46{\scriptstyle\pm.03}$ \\
\addlinespace
L Post.\ Oblique & T & $-.02{\scriptstyle\pm.05}$ & $.37{\scriptstyle\pm.03}$ & $-.03{\scriptstyle\pm.08}$ & $.21{\scriptstyle\pm.03}$ & $\mathbf{.17}{\scriptstyle\pm.10}$ & $\mathbf{.44}{\scriptstyle\pm.04}$ \\
                 & W & $-.04{\scriptstyle\pm.04}$ & $.37{\scriptstyle\pm.04}$ & $-.08{\scriptstyle\pm.06}$ & $.19{\scriptstyle\pm.05}$ & $.10{\scriptstyle\pm.08}$ & $.45{\scriptstyle\pm.02}$ \\
\addlinespace
Anterior         & T & $-.04{\scriptstyle\pm.01}$ & $-.00{\scriptstyle\pm.06}$ & $-.07{\scriptstyle\pm.02}$ & $.14{\scriptstyle\pm.05}$ & $\mathbf{.00}{\scriptstyle\pm.04}$ & $\mathbf{.37}{\scriptstyle\pm.04}$ \\
                 & W & $-.04{\scriptstyle\pm.02}$ & $.02{\scriptstyle\pm.06}$ & $-.08{\scriptstyle\pm.03}$ & $.18{\scriptstyle\pm.04}$ & $-.04{\scriptstyle\pm.03}$ & $.40{\scriptstyle\pm.04}$ \\
\addlinespace
R Ant.\ Oblique  & T & $-.06{\scriptstyle\pm.01}$ & $.22{\scriptstyle\pm.05}$ & $-.09{\scriptstyle\pm.02}$ & $.10{\scriptstyle\pm.02}$ & $-.06{\scriptstyle\pm.03}$ & $.31{\scriptstyle\pm.06}$ \\
                 & W & $-.06{\scriptstyle\pm.01}$ & $.17{\scriptstyle\pm.07}$ & $-.13{\scriptstyle\pm.07}$ & $.15{\scriptstyle\pm.04}$ & $-.10{\scriptstyle\pm.03}$ & $.32{\scriptstyle\pm.06}$ \\
\addlinespace
Posterior        & T & $-.03{\scriptstyle\pm.00}$ & $-.11{\scriptstyle\pm.03}$ & $-.05{\scriptstyle\pm.02}$ & $-.00{\scriptstyle\pm.06}$ & $-.05{\scriptstyle\pm.02}$ & $\mathbf{.27}{\scriptstyle\pm.03}$ \\
                 & W & $-.03{\scriptstyle\pm.00}$ & $-.08{\scriptstyle\pm.01}$ & $-.07{\scriptstyle\pm.03}$ & $.00{\scriptstyle\pm.04}$ & $-.10{\scriptstyle\pm.03}$ & $.21{\scriptstyle\pm.01}$ \\
\bottomrule
\end{tabular*}
\end{table*}

\subsection{Sagittal-View Z-Score Regression}

\begin{figure*}[!t]
  \centering
  \includegraphics[width=\textwidth]{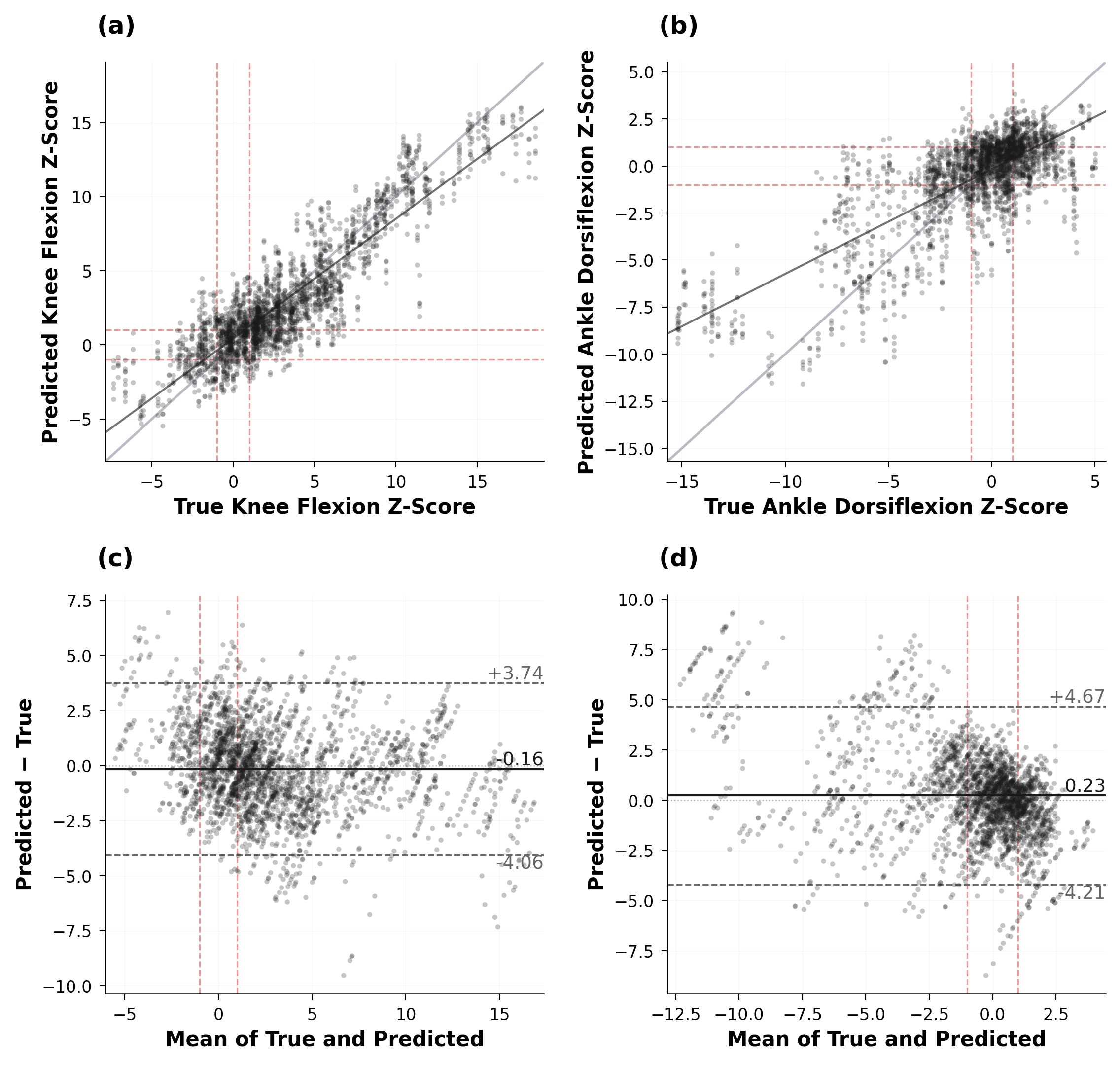}
  \caption{Trial-level z-score prediction on AGCN+ViT. Top row, predicted vs.\ true z-scores for (a) knee and (b) ankle. The black line is the regression fit and the gray line is the identity. Red dashed lines mark $\pm 1$ classification boundaries. Bottom row, Bland-Altman plots for (c) knee (bias $= -0.16$, LoA $= [-4.06, 3.74]$) and (d) ankle (bias $= 0.23$, LoA $= [-4.21, 4.67]$).}
  \label{fig:scatter_ba}
\end{figure*}

Table~\ref{tab:regression_detailed} reports the trial level performance for the two best models (AGCN~\citep{shi_two_2019} with average pooling and AGCN+ViT~\citep{dosovitskiy_image_2021} hierarchical aggregation) on the sagittal view.
AGCN~\citep{shi_two_2019}+ViT~\citep{dosovitskiy_image_2021} achieves $R^2 = 0.80 \pm 0.02$ and a concordance correlation coefficient (CCC) of $0.89 \pm 0.02$ for knee z-scores.
Ankle z-score prediction is lower ($R^2 = 0.57 \pm 0.02$, CCC $= 0.72 \pm 0.02$), with a positive bias of $0.23 \pm 0.14$ indicating slight overestimation toward dorsiflexion.
All models predict knee z-scores more accurately than ankle z-scores.
\autoref{fig:scatter_ba} shows predicted-versus-true scatter plots and Bland-Altman agreement analysis for the ViT~\citep{dosovitskiy_image_2021} model at trial level.
Knee predictions (a) track the identity line across the full range ($z \in [-9, 20]$), whereas ankle predictions (b) systematically underestimate severity for extreme plantarflexion values ($z < -10$).
Bland-Altman analysis shows near-zero bias for both joints (knee: $-0.16$, ankle: $0.23$), with knee limits of agreement (LoA) of $[-4.06, 3.74]$ (c) and wider ankle LoA of $[-4.21, 4.67]$ (d), with errors increasing for extreme values.

\begin{table}[!t]
\centering
\caption{Detailed regression metrics for AGCN and AGCN+ViT on the sagittal view. All results report mean $\pm$ 95\% CI across 5 iterations.}
\label{tab:regression_detailed}
\small
\begin{tabular}{llccccc}
\toprule
Joint & Model & $R^2$ & MAE & RMSE & CCC & Bias \\
\midrule
\multirow{2}{*}{Knee}
& AGCN & $.77 \pm .01$ & $1.65 \pm 0.04$ & $2.13 \pm 0.04$ & $.87 \pm .01$ & $-.25 \pm .13$ \\
& +ViT & $\mathbf{.80} \pm .02$ & $\mathbf{1.55} \pm 0.07$ & $\mathbf{2.00} \pm 0.10$ & $\mathbf{.89} \pm .02$ & $-.16 \pm .14$ \\
\midrule
\multirow{2}{*}{Ankle}
& AGCN & $.57 \pm .04$ & $1.63 \pm 0.07$ & $2.28 \pm 0.12$ & $.71 \pm .03$ & $.44 \pm .13$ \\
& +ViT & $.57 \pm .02$ & $1.64 \pm 0.05$ & $2.28 \pm 0.07$ & $\mathbf{.72} \pm .02$ & $.23 \pm .14$ \\
\bottomrule
\end{tabular}
\end{table}

\subsection{Rodda and Graham 7-Class Gait Classification}

\begin{figure}[!t]
  \centering
  \includegraphics[width=\columnwidth]{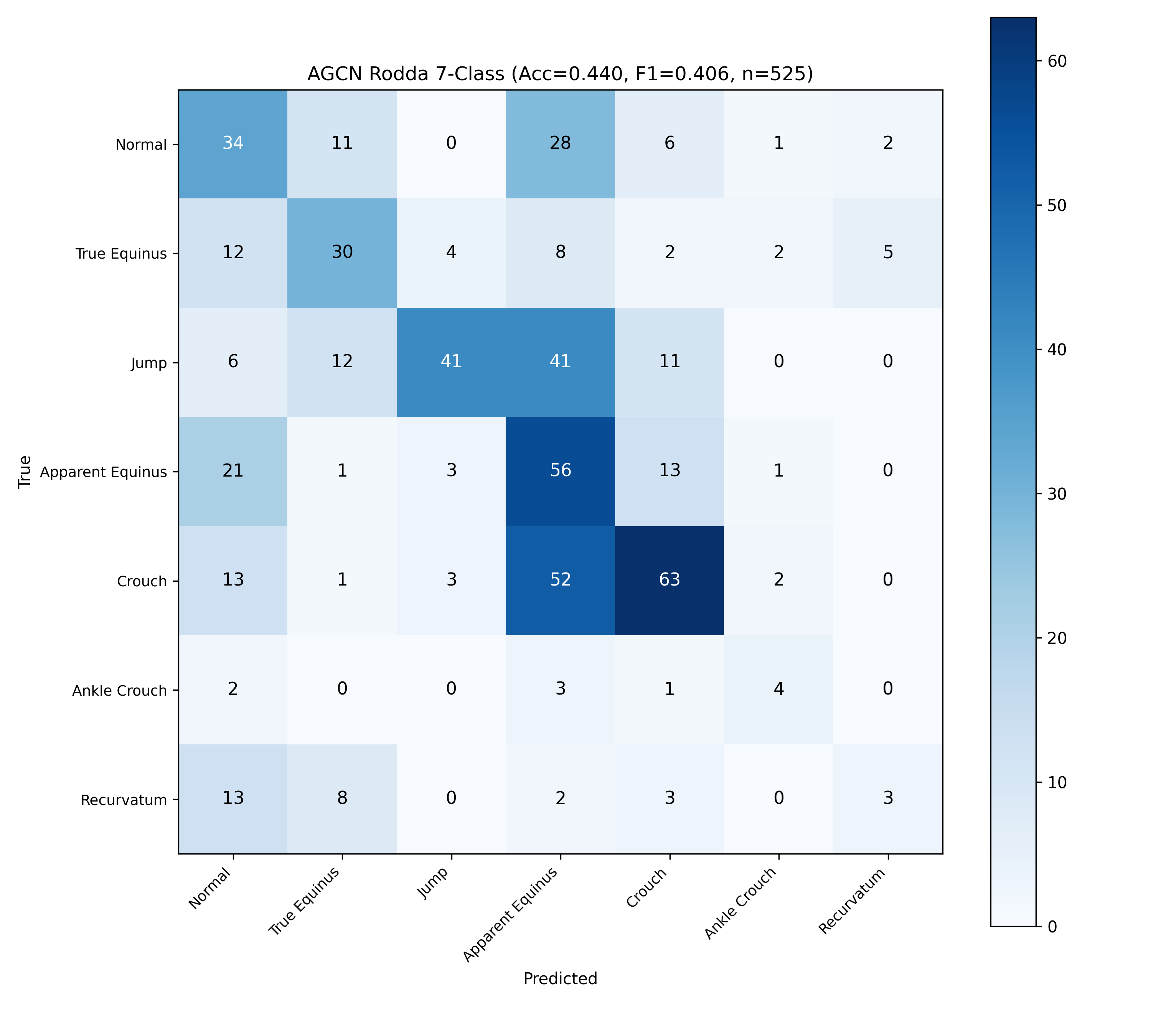}
  \caption{Confusion matrix for Rodda and Graham 7-class classification from predicted z-scores (AGCN).}
  \label{fig:rodda_confusion}
\end{figure}

We applied the Rodda and Graham classification rules (\autoref{tab:rodda_regions}) to the predicted ankle and knee z-scores to classify each trial into one of seven gait patterns.
Using AGCN+ViT trial-level predictions, the 7-class accuracy is $43 \pm 1\%$ with macro-F1 $= 0.37 \pm 0.03$ and macro-AUROC $= 0.78 \pm 0.01$.
Table~\ref{tab:rodda_per_class} reports per-class metrics.
Crouch gait (F1 $= 0.53$, $n = 134$) and Jump gait (F1 $= 0.51$, $n = 111$) are the most reliably identified classifications.
The confusion matrix (\autoref{fig:rodda_confusion}) shows that Jump and Crouch trials are frequently misclassified as Apparent Equinus.
Recurvatum achieves the lowest F1 ($0.15$, recall $= 11\%$, $n = 29$).

\begin{table}[!t]
\centering
\caption{Rodda and Graham 7-class per-class metrics (one-vs-rest, AGCN, mean across 5 iterations). Overall accuracy $= 43 \pm 1\%$, macro-F1 $= 0.37 \pm 0.03$, macro-AUROC $= 0.78 \pm 0.01$.}
\label{tab:rodda_per_class}
\small
\begin{tabular}{lrccccc}
\toprule
Class & $n$ & F1 & Prec & Rec & AUROC & AUPRC \\
\midrule
Normal          & 83  & .38 & .34 & .44 & .77 & .32 \\
True Equinus    & 63  & .45 & .46 & .45 & .84 & .46 \\
Jump            & 111 & .51 & .73 & .39 & .76 & .60 \\
Apparent Equinus & 95 & .36 & .28 & .51 & .68 & .28 \\
Crouch          & 134 & .53 & .61 & .46 & .79 & .61 \\
Ankle Crouch    & 10  & .22 & .22 & .22 & .77 & .08 \\
Recurvatum      & 29  & .15 & .24 & .11 & .82 & .17 \\
\bottomrule
\end{tabular}
\end{table}

\subsection{Binary Classification of Excess Knee Flexion}

\begin{figure}[!t]
  \centering
  \includegraphics[width=\columnwidth]{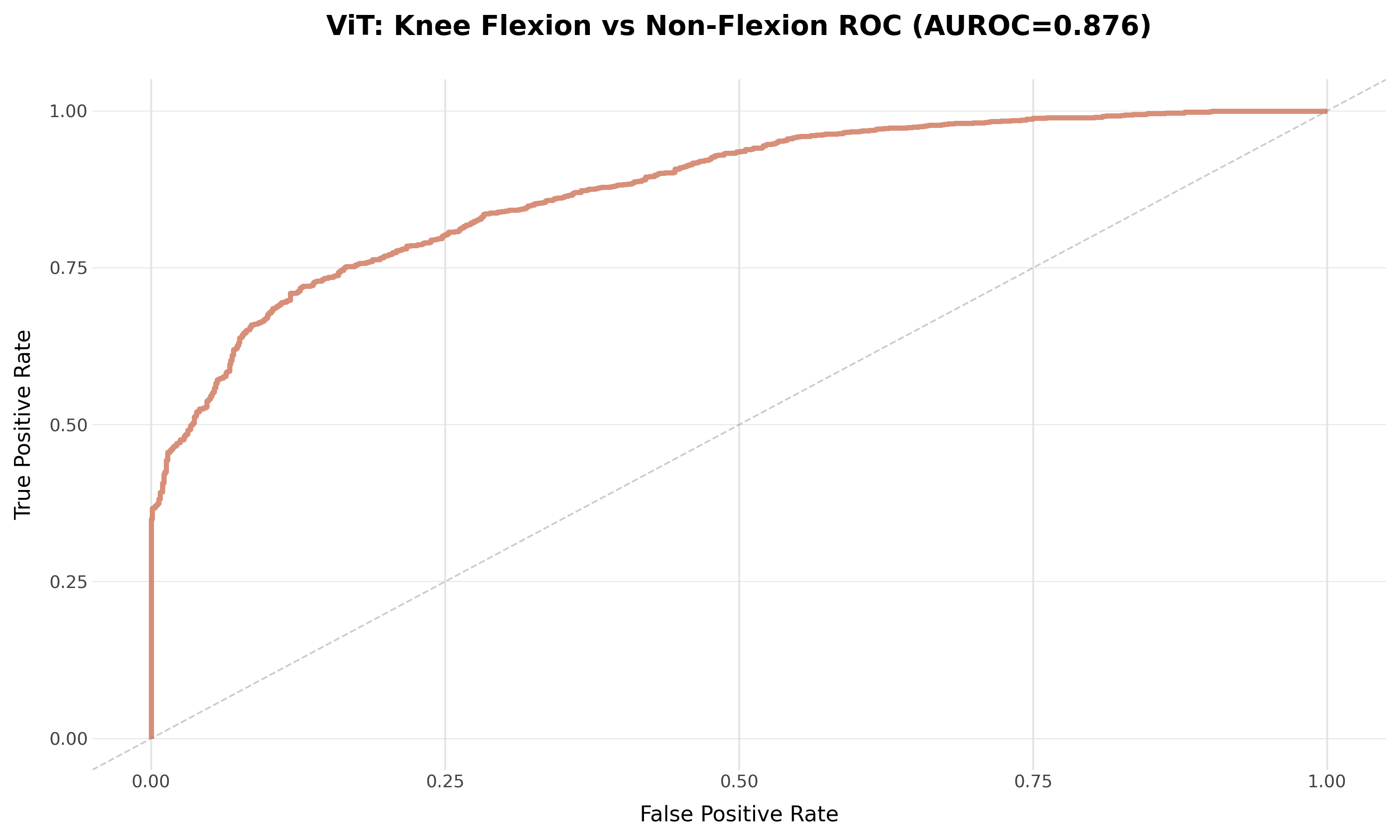}
  \caption{ROC curve for binary knee flexion classification (AGCN+ViT). AUROC $= 0.88$.}
  \label{fig:roc_flexion}
\end{figure}

Excess knee flexion (crouch gait) is the most prevalent sagittal-plane gait deviation in ambulatory children with CP, affecting more than half of GMFCS I to III children and rising in prevalence with age~\citep{rethlefsen_prevalence_2017}.
The clinical decision threshold for this pattern is a single cut-point on the knee z-score ($z > 1$), making binary detection of $z > 1$ the most natural first test of whether our continuous z-score predictions can support a clinically useful triage decision.
Thresholding predicted knee z-scores at $+1$ yielded a binary classifier for excess knee flexion.
Table~\ref{tab:knee_flexion} reports classification metrics.
The AGCN+ViT~\citep{dosovitskiy_image_2021} model achieves AUROC $= 0.88 \pm 0.02$ and AUPRC $= 0.93 \pm 0.01$.
Recall is $0.83 \pm 0.02$ and specificity is $0.72 \pm 0.01$, corresponding to a 28\% false-positive rate among non-flexion cases (\autoref{fig:roc_flexion}).

\begin{table}[!t]
\centering
\caption{Binary classification of excess knee flexion ($z > 1$ vs.\ $z \leq 1$). Mean $\pm$ 95\% CI across 5 iterations.}
\label{tab:knee_flexion}
\footnotesize
\setlength{\tabcolsep}{3pt}
\begin{tabular}{lccccccc}
\toprule
Model & Acc & F1 & Prec & Rec & Spec & AUROC & AUPRC \\
\midrule
AGCN & $.78{\scriptstyle\pm.01}$ & $.83{\scriptstyle\pm.00}$ & $.84{\scriptstyle\pm.02}$ & $.82{\scriptstyle\pm.02}$ & $.71{\scriptstyle\pm.05}$ & $.86{\scriptstyle\pm.01}$ & $.92{\scriptstyle\pm.01}$ \\
+ViT & $\mathbf{.79}{\scriptstyle\pm.02}$ & $\mathbf{.84}{\scriptstyle\pm.02}$ & $\mathbf{.84}{\scriptstyle\pm.01}$ & $\mathbf{.83}{\scriptstyle\pm.02}$ & $\mathbf{.72}{\scriptstyle\pm.01}$ & $\mathbf{.88}{\scriptstyle\pm.01}$ & $\mathbf{.93}{\scriptstyle\pm.01}$ \\
\bottomrule
\end{tabular}
\end{table}

\subsection{Error Stratification by Classification Difficulty}

Aggregate $R^2$ and CCC do not reveal whether the model fails uniformly across the z-score range or instead concentrates its errors near the $\pm 1$ classification boundaries, the region where downstream class assignment is most sensitive to small prediction errors and where observational clinicians themselves show the lowest inter-rater agreement~\citep{kim_reliability_2011}.
We therefore profiled the model for clinical use by stratifying trials according to their distance from the nearest boundary, identifying the z-score regions in which the model's uncertainty is small enough to support a clinical decision and the regions in which it is not.
To assess whether regression accuracy varies with proximity to the $\pm 1$ classification boundaries, we stratified knee trials by the distance of their true z-score $z$ from the nearest boundary into \emph{hard} cases, defined as $ 0.5\leq |z| \leq 1.5$, i.e., within $0.5$ units of either the +1 or -1 boundary, and \emph{easy} cases, defined as the complement $|z| \in [0, 0.5] \cup [1.5, \infty)$.
Table~\ref{tab:easy_hard} reports the results.
Easy trials retain strong agreement (ViT~\citep{dosovitskiy_image_2021} CCC $= 0.89$, $R^2 = 0.80$), whereas hard trials yield CCC $= 0.27$ and $R^2 = -1.60$.

\begin{table}[!t]
\centering
\caption{Knee z-score regression stratified by distance of the true z-score $z$ from the nearest $\pm 1$ classification boundary. Hard  is $|z| \in (0.5, 1.5)$ (within $0.5$ units of a boundary). Easy is the complement, $|z| \in [0, 0.5] \cup [1.5, \infty)$. Mean [95\% CI] across 5 iterations.}
\label{tab:easy_hard}
\small
\begin{tabular}{llcccc}
\toprule
Model & Stratum & $R^2$ & MAE & CCC & Acc \\
\midrule
\multirow{2}{*}{AGCN}
& Easy & $.77 \pm .01$ & $1.78 \pm 0.05$ & $.87 \pm .01$ & $.75 \pm .01$ \\
& Hard & $-1.36 \pm 0.27$ & $1.13 \pm 0.06$ & $.24 \pm .02$ & $.50 \pm .01$ \\
\addlinespace
\multirow{2}{*}{+ViT}
& Easy & $.80 \pm .02$ & $1.63 \pm 0.10$ & $.89 \pm .02$ & $.76 \pm .02$ \\
& Hard & $-1.60 \pm 0.40$ & $1.23 \pm 0.09$ & $.27 \pm .04$ & $.49 \pm .06$ \\
\bottomrule
\end{tabular}
\end{table}

\begin{figure*}[!t]
  \centering
  \includegraphics[width=\textwidth]{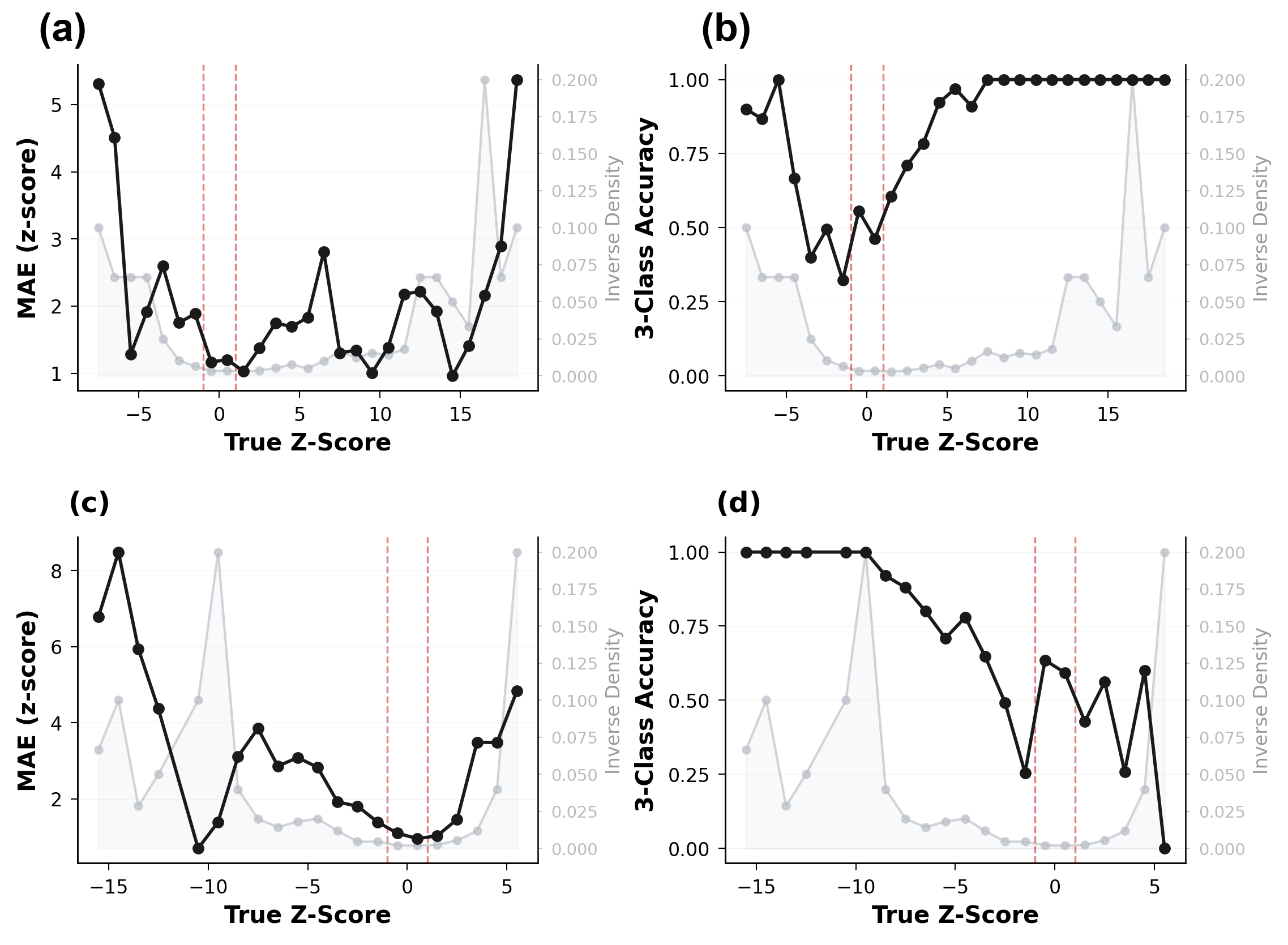}
  \caption{Per-bin analysis of trial-level AGCN+ViT z-score predictions. The true z-score range is partitioned into contiguous bins of width $0.5$, and mean absolute error (MAE) and 3-class accuracy are computed within each bin. (a) Knee per-bin MAE with label-distribution-smoothed inverse sample density overlay (gray). (b) Knee per-bin 3-class accuracy. (c) Ankle per-bin MAE with inverse sample density overlay. (d) Ankle per-bin 3-class accuracy. 3-class assignment uses $z < -1$, $-1 \leq z \leq 1$, $z > 1$. Red dashed lines mark $\pm 1$ boundaries. Bins with fewer than $3$ samples are excluded.}
  \label{fig:per_bin}
\end{figure*}

To examine how prediction quality varies with true z-score magnitude, we partitioned the true z-score range into contiguous bins of width $0.5$ and computed mean absolute error (MAE) and 3-class accuracy within each bin, treating any bin with fewer than 3 trials as unreliable and excluding it from the analysis.
\autoref{fig:per_bin} shows the resulting per-bin metrics separately for the knee (\autoref{fig:per_bin}a, b) and ankle (\autoref{fig:per_bin}c, d) under AGCN+ViT trial-level prediction.
For the knee (\autoref{fig:per_bin}a, b), the bins nearest the $\pm 1$ boundaries contain the highest sample density ($n = 290$--$345$) and achieve the lowest per-bin MAE ($\approx 1.0$), yet 3-class accuracy in these bins drops to 45--55\%.
In contrast, bins far from any boundary ($z > 5$) achieve near-perfect 3-class accuracy (95--100\%) despite substantially higher MAE ($2$--$5$).
Bins at the extremes ($z > 15$) contain very few trials ($n = 5$--$15$) and exhibit the highest MAE.
The label-distribution-smoothed (LDS) inverse sample density overlay in \autoref{fig:per_bin}a (gray curve), a measure of how underrepresented each z-score region is in the training data~\citep{yang_delving_2021}, shows that higher z-score magnitudes correspond to sparser sample density, and the per-bin MAE is strongly positively correlated with this sparsity (Pearson $r = 0.715$, $p < 0.001$).
The ankle joint (\autoref{fig:per_bin}c, d) follows the same pattern in amplified form: the densest bins near $z = 0$ ($n = 106$--$115$) have the lowest MAE ($\approx 0.8$--$1.0$) but the worst 3-class accuracy (45--55\%), while extreme plantarflexion bins ($z < -10$, $n = 3$--$6$) achieve 100\% classification accuracy despite MAE of $4$--$8$.
On the dorsiflexion side ($z > 1$), moderate sample sizes ($n = 11$--$51$) yield low accuracy (25--55\%).
The correlation between inverse density and MAE is even stronger for ankle ($r = 0.887$, $p < 0.001$).

\subsection{Systematic Underestimation of Gait Severity}

\begin{figure*}[!t]
  \centering
  \includegraphics[width=0.48\textwidth]{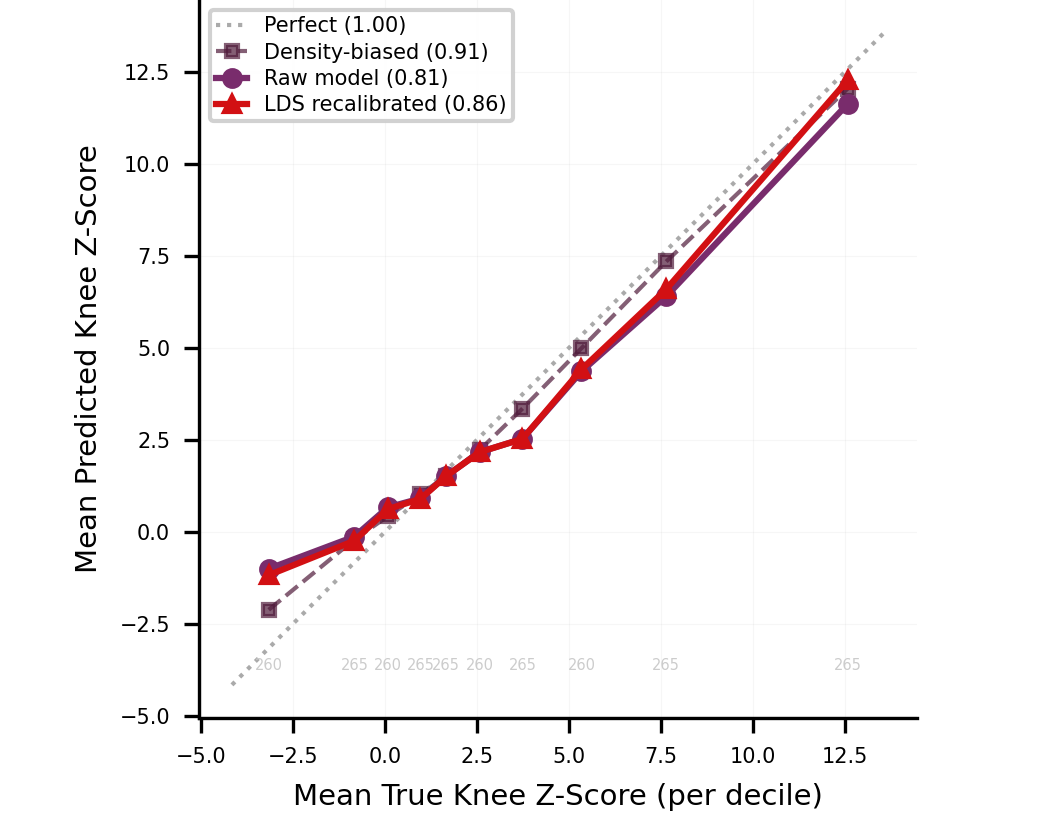}%
  \hfill
  \includegraphics[width=0.48\textwidth]{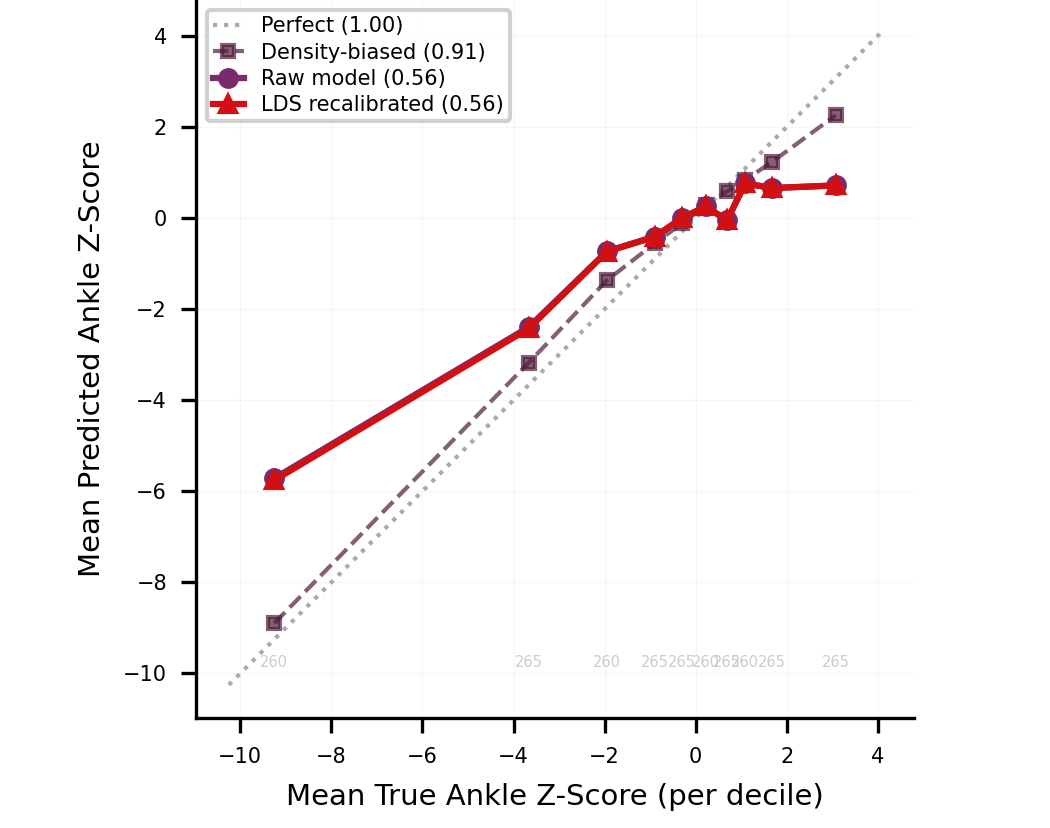}
  \caption{Decile-binned calibration plots for knee (left) and ankle (right). The model line (colored) shows mean predicted vs.\ mean true z-score per decile, where a perfectly calibrated model would follow the identity (dotted).}
  \label{fig:calibration}
\end{figure*}

The z-score label distribution in our cohort is highly imbalanced, with the majority of trials concentrated near zero and the most clinically severe cases underrepresented in the tails, a regime in which deep regression models are known to systematically underestimate the magnitude of extreme targets~\citep{yang_delving_2021}.
We therefore profiled this bias using decile-binned calibration analysis~\citep{steyerberg_assessing_2010} and tested whether a label-distribution-aware recalibration can correct it, since a systematic and correctable bias has direct implications for whether the predictor can be deployed without modification on the very cases that drive treatment decisions.
To quantify the degree to which the AGCN+ViT model underpredicts severity, we computed decile-binned calibration slopes for both joints~\citep{steyerberg_assessing_2010} (\autoref{fig:calibration}).
This involved sorting trials into ten equal-population bins by true z-score and plotted the per-bin mean predicted value against per-bin mean true value, so a slope of $1$ means the model recovers the full severity range and a slope below $1$ means it systematically compresses extreme predictions toward the population mean.
A perfectly calibrated model would produce a slope of $1.0$ with flatter slopes indicating greater systematic underestimation of deviation from typical gait.
The AGCN+ViT model trained on Knee achieves a calibration slope of $0.81$, while AGCN+ViT trained on Ankle is $0.56$.
A sample density biased perfect predictor that predicts each trial's z-score as the Gaussian-weighted local mean of the training distribution yielded slopes of $0.91$ for both joints.
In plain terms, this is the calibration slope an idealized predictor would still achieve even if its only error came from training-set imbalance, with $0.91$ marking the compression of severity attributable to the data being densely concentrated near zero and sparse at the extremes, against which any real model's slope can be compared to isolate model-side bias.
The excess underestimation beyond the density biased perfect predictor is therefore $-0.35$ for ankle and $-0.10$ for knee.
We attempt recalibrating the raw baseline models using LDS, in which the binned label count over a $0.5$ z-score interval is smoothed with a Gaussian kernel.
This adjusts model predictions based on the known training-set label density in the z-score space, biasing predictions from denser regions toward lower-density z-score regions.
This counteracts the model's drift toward the dense middle of the label distribution by nudging outputs back toward the rare extremes.
After recalibration, the bias relative to the density biased perfect predictor persists unchanged in ankle ($-0.35$) and is reduced in knee ($-0.05$).

\section{Discussion}
\label{sec:discussion}

\subsection{Direct Biomechanical Calculation From Monocular Video Fails}

The Rodda and Graham z-score is, by construction, a deterministic function of sagittal-plane joint angles, and we calculate the knee and ankle angles which are the required inputs to produce Z-scores.
A biomechanical baseline without machine learning that simply applies the z-score formula (\autoref{eq:zscore}) to the monocular-derived 3D angles, therefore probes whether learning is required at all.
Every viewpoint yielded negative $R^2$ for both joints, with the sagittal view (knee $R^2 = -1.30$, ankle $R^2 = -0.38$) only marginally worse than predicting the population mean and the posterior view (knee $R^2 = -4.51$, ankle $R^2 = -1.11$) substantially worse (\autoref{tab:multiview_angles}, Biomech column).
This collapse cannot be attributed to the z-score formulation itself, since the same arithmetic applied to 3D-IGA angles defines the clinical gold standard.
Joint-angle estimates from markerless pose estimation carry several-degree errors even in multi-camera systems, with reported sagittal-plane RMSE around 3 to 6 degrees at the lower-extremity joints relative to marker-based 3D-IGA in concurrent comparisons~\citep{kanko_concurrent_2021}, and monocular pose estimators amplify these errors further by removing the multi-view geometric constraints that bound out-of-plane reconstruction~\citep{cronin_using_2021}.
Angular errors of this magnitude are large enough  to cause significant Z-score variation when directly calculated.
Sagittal viewpoints partially preserve in-plane geometry and degrade least. 
Other views fail completely perhaps due to lack of information in those views to identify accurate knee and ankle angles (\autoref{tab:multiview_angles}, Biomech column).

The implication is twofold.
First, current monocular markerless motion capture is not yet accurate enough to support direct biomechanical computation of clinical metrics like Rodda and Graham z-scores. 
Published video-based methods that report strong agreement with marker-based gold standards typically do so on coarse spatiotemporal parameters such as cadence and walking speed~\citep{azhand_algorithm_2021, kidzinski_deep_2020} rather than on per-frame joint angles, and our results suggest that the accuracy of outputs of monocular pose estimation systems are unsuitable for traditional biomechanical analysis.
Second, the role of the deep learning models evaluated below is precisely to denoise and reweight this noisy angular signal, learning the residual mapping from monocular-estimate angles to 3D-IGA-equivalent z-scores.
The gap between the Biomech baseline ($R^2 < 0$ on the best-performing sagittal view, \autoref{tab:multiview_angles}) and trial-level AGCN+ViT (knee $R^2 = 0.80$, ankle $R^2 = 0.57$, \autoref{tab:regression_detailed}) quantifies how much z-score signal is recoverable from monocular angles only by exploiting the cross-frame, cross-joint, and cross-cycle structure that direct calculation discards.

\subsection{Regression Performance and Clinical Benchmarks}

The three architectures differ fundamentally in how they represent joint relationships, and this predicts their relative performance on z-score regression.
AGCN's~\citep{shi_two_2019} attention mechanism learns a data-driven adjacency matrix that adapts joint connectivity on a per-sample basis (Section~\ref{sec:models}, \nameref{sec:models}), capturing person-specific relationships beyond the predefined skeletal topology enabling it to achieve a trial-level $R^2 = 0.77 \pm 0.01$ for knee and $R^2 = 0.57 \pm 0.04$ for ankle under average-pool aggregation (\autoref{tab:regression_detailed}).
Gait deviations in CP form a heterogeneous continuum rather than discrete categories~\citep{armand_gait_2016}. 
Identical Rodda and Graham classifications can arise from distinct motion kinematics in body parts outside the knee and ankle~\citep{pandey_crouch_2023}.
AGCN~\citep{shi_two_2019} addresses this heterogeneity directly by learning per-sample graph topology, whereas ST-GCN~\citep{yan_spatial_2018}, constrained to fixed skeletal edges, cannot adapt to inter-patient variability and underperforms both alternatives across all eight camera viewpoints (\autoref{tab:multiview_angles}).
The results show that DCL~\citep{ordonez_deep_2016} which carries no skeletal graph structure at all ranks second on both targets (\autoref{tab:multiview_angles}), indicating that long range temporal dynamics aggregated by the LSTM within the joint angle sequences carry strong predictive signal that the ST-GCN lacks in its temporal graph convolutions.
Hierarchical aggregation via  AGCN+ViT~\citep{dosovitskiy_image_2021} improves the trial-level knee $R^2$ from $0.77$ (AGCN with average pooling) to $0.80 \pm 0.02$ (AGCN+ViT) and CCC from $0.87$ to $0.89 \pm 0.02$ (\autoref{tab:regression_detailed}).
This may be due to naive average pooling diluting high-quality cycle predictions with uninformative segments, whereas ViT hierarchical aggregation learns to down-weight uninformative cycles.
For ankle Z-score regression, AGCN+ViT~\citep{dosovitskiy_image_2021} does not improve $R^2$ compared to AGCN only modeling, though CCC increases marginally from $0.71$ to $0.72$ (\autoref{tab:regression_detailed}).
Despite the improvements to knee z-score regression due to hierarchical aggregation in AGCN+ViT, the Bland-Altman analysis (\autoref{fig:scatter_ba}c, d) reveals that the residual variance is strongly dependent on the relative density of samples in the Z-score region, with higher variance in regions with fewer samples which in our case was the high severity extremes >2 standard deviations away from 0. 
In the normative z-score region of each joint distribution where the mean of true and predicted lies within $[-1, 1]$, per-trial errors cluster within roughly $\pm 1$ z-score unit of the bias line, whereas toward the tails the spread expands by 3- to 5-fold.
The residuals increase with severity, indicating that the model regresses extreme predictions toward the population mean.
At high knee severity (mean of true and predicted $> 10$), the predicted-true differences skew negative.
Similarly, at extreme ankle plantarflexion (mean $< -10$) they skew strongly positive.
In both cases the model under-predicts the magnitude of the true deviation, the same severity underestimation that the calibration analysis directly quantifies (\autoref{fig:calibration}).

\subsection{Boundary Effects and the Limits of Regression-Based Classification}

Regression accuracy collapses at classification boundaries. Under AGCN+ViT trial-level prediction, easy trials ($\geq 0.5$ z-score units from any $\pm 1$ threshold) achieve CCC $= 0.89$, whereas hard trials ($< 0.5$ units) yield CCC $= 0.27$ and $R^2 = -1.60$, worse than the overall baseline (\autoref{tab:easy_hard}).
This loss in performance in the hard region follows directly from the AGCN+ViT model's MAE of $1.55$ (\autoref{tab:regression_detailed}) which makes boundary-region classification unreliable, especially considering that the entire normative region is just 2.0 units wide.
Per-bin analysis supports this finding (\autoref{fig:per_bin}).
Boundary-adjacent bins contain the highest sample density and the lowest per-bin MAE ($\approx 1.0$ for knee, $\approx 0.8$ for ankle), yet classification accuracy in this range is lowest (45--55\%), because even a modest absolute error of 1.0 units misclassifies cases that are similar in severity to TD groups.
Conversely, bins far from any boundary achieve near-perfect classification despite 2--5$\times$ higher MAE, because large true deviations tolerate proportionally large prediction errors without crossing a category threshold.
Extreme-range bins compound this problem through data imbalance. 
Bins at the tails of the z-score distribution contain only $n = 3$--$15$ samples and exhibit the highest per-bin MAE.
Per-bin MAE correlates strongly with LDS-smoothed inverse sample density (knee $r = 0.715$, ankle $r = 0.887$, both $p < 0.001$, \autoref{fig:per_bin}a, c), consistent with the established degradation of deep regression models in underrepresented target regions~\citep{yang_delving_2021}.
Boundary-region errors carry clinical consequence beyond their statistical magnitude.
A child whose true knee z-score is $1.1$ but predicted as $0.4$ would be classified as kinematically normal, postponing the precise quantification that informs whether to consider surgical hamstring lengthening or non-operative monitoring, while a child whose true z-score is $0.9$ predicted as $1.4$ would be referred for an unnecessary treatment workup.
Closing this gap requires reducing per-trial MAE in the boundary region to below approximately $0.5$ z-score units, half the width of the normative band.
Our results suggest that simply scaling sample size in this region is unlikely to deliver this reduction, since the boundary-adjacent bins already contain the highest sample density in our cohort yet still produce per-bin MAE near $1.0$ (\autoref{fig:per_bin}b, d).
The remaining error is therefore methodological, in that it reflects the limited ability of current pipelines to resolve fine-grained kinematic differences rather than a labeled-data shortage.

Two architectural directions are likely necessary.
First, the monocular markerless pose estimator must capture sub-degree distinctions in joint angles that the current MoVi-trained backbone~\citep{ghorbani_movi_2021} treats as equivalent under its normative motion prior, an upstream limitation that no downstream regressor can fully recover.
Second, the skeleton-based architectures evaluated here are inherited from human action recognition, where training targets distinguish gross motor patterns such as walking from running rather than the subtle within-class variations that separate a stride with mild excessive flexion from one within the normative band.
The same limitation has been documented in fine-grained settings, where Human action recognition models trained on coarse action labels struggle to detect the kinematically subtle but task-relevant differences between visually similar manipulations~\citep{wang_chefvl_2025}.

\subsection{Ankle Prediction as the Primary Bottleneck}

Ankle z-score error emerges as the primary bottleneck at three levels: lower regression agreement (CCC $= 0.72$ vs.\ $0.89$ for knee, \autoref{tab:regression_detailed} and \autoref{fig:scatter_ba}c, d), systematic underestimation of severity for extreme plantarflexion values (\autoref{fig:per_bin}c, d and \autoref{fig:calibration}), and predominant contribution to misclassification in Rodda and Graham classification (\autoref{tab:rodda_per_class}, \autoref{fig:rodda_confusion}).
Three factors explain the performance gap between ankle and knee.
First, the knee has a larger sagittal-plane range of motion than the ankle, making its flexion and extension patterns more discriminable from 2D video.
Second, standard motion capture datasets such as MoVi~\citep{ghorbani_movi_2021} consist exclusively of typically developing adults performing everyday actions. 
CP-specific ankle movement patterns, such as equinus and severe plantarflexion, are absent from this normative population, and this underrepresentation likely contributes to noisier ankle and foot keypoints when pose estimators trained on such data are applied to CP gait, consistent with the established finding that pose-estimation accuracy degrades sharply on body configurations and populations underrepresented in the training distribution~\citep{cronin_using_2021}.
Third, while the ankle z-score distribution is wide (range $[-19, +5]$), most labels are near zero and causing the model to systematically underestimate severity for the most impaired children.
Calibration analysis isolates a pose-estimation-specific contribution to this severity underestimation, beyond what sample density alone predicts (\autoref{fig:calibration}).
The raw calibration slopes corroborate this asymmetry, with knee predictions capturing 81\% of the true z-score range while ankle predictions compress to roughly half (slope $0.56$).
The ankle model's excess underestimation beyond the density baseline ($-0.35$) is approximately thrice that of the knee model ($-0.10$), despite both joints sharing the same density biased perfect predictor slope ($0.91$).
LDS applied to the trained models brings the knee excess underestimation down to $-0.05$ but leaves the ankle excess unchanged at $-0.35$, evidence that the residual ankle bias arises outside the sample-density mechanism that LDS targets and therefore likely upstream in the pose estimator.
This disparity implicates the pose estimator's training distribution: because equinus and severe plantarflexion are rare in normative gait datasets, the pose estimator defaults ankle keypoints toward typical positions when confronted with these configurations, whereas atypical knee flexion is well-represented in everyday movements such as running and stair climbing.
The calibration curves from the trained models support our previous discussion that the pose estimator has a systematic error, where it reduces ankle range of motion to more normative range. 

\subsection{From Z-Score Regression to Gait Pattern Classification}

Predicted z-scores can be used for downstream Rodda and Graham classification only when the severity is high.
Trial-level binary detection of excess knee flexion ($z > 1$) under AGCN+ViT achieves AUROC $= 0.88 \pm 0.02$ and recall $= 0.83$ (\autoref{tab:knee_flexion}, \autoref{fig:roc_flexion}).

Apparent Equinus is the dominant confusion target under AGCN trial-level predictions because it occupies the $|z_\text{ankle}| < 1$ region (\autoref{fig:rodda_confusion}). It shares the knee $z > 1$ criterion with Crouch and Jump but is distinguished solely by ankle z-score.
Because the model systematically underestimates ankle deviation, trials with true ankle z-scores near $\pm 1$ are predicted as less severe and misclasssified into the Apparent Equinus region.
The confusion matrix (\autoref{fig:rodda_confusion}) shows the misclassification, with $33\%$ of all true Jump trials and $38\%$ of all true Crouch trials predicted as Apparent Equinus.
Apparent Equinus has the lowest per-class AUROC of the four primary Rodda and Graham classifications ($0.68$, compared to $0.84$ for True Equinus, $0.79$ for Crouch, and $0.76$ for Jump), corroborating that Apparent Equinus operates as a default sink for ankle-underestimated trials rather than as a class with its own learnable signature.
The fact that True Equinus, Crouch, and Jump retain AUROC values between $0.76$ and $0.84$ indicates that the model has ability to rank severity of these conditions, which supports the longitudinal trajectory-tracking deployment introduced above where relative change in z-score matters more than hard category assignment.
Recurvatum is nearly undetectable (F1 $= 0.15$, recall $= 10\%$, \autoref{tab:rodda_per_class}). 
With only $n = 29$ limbs, the model lacks sufficient training samples to distinguish this rare class and the pose estimator is unlikely to have experiences recurvatum gait during training.

\subsection{Clinical Relevance}

The most consequential clinical use of continuous z-score regression is longitudinal monitoring rather than cross-sectional classification.
Tracking ankle and knee z-scores extracted from routine clinic video across visits provides an objective, quantitative substrate for detecting disease progression and treatment response, a capability that observational rating scales fundamentally cannot provide at the within-person sensitivity required for clinical decision-making.
For this use case, continuous regression output is inherently better suited than categorical labeling, since trajectory change is detectable in z-score units long before a child crosses a $\pm 1$ category threshold.
Whether the per-trial uncertainty quantified above is small relative to the within-person change typically observed across treatment intervals such as botulinum toxin injections, hamstring lengthening, and orthotic modification is the central remaining question for prospective validation.

Trial-level binary knee flexion screening using AGCN+ViT (AUROC $= 0.88$, \autoref{tab:knee_flexion} and \autoref{fig:roc_flexion}) is the system's most deployable option in the near term.
In a community clinic, the model would function as a triage instrument where children flagged with excess knee flexion ($z > 1$) are referred for full 3D-IGA, while those below threshold are monitored with video-based assessment alone.
At 28\% false-positive rate (specificity $= 0.72$, \autoref{tab:knee_flexion}), over-referral risk is low. Missing progressive crouch gait carries greater clinical consequence than an unnecessary 3D-IGA appointment.
Children near the classification boundary are precisely those for whom clinical observational assessment also shows the lowest inter-rater agreement~\citep{kim_reliability_2011}, with reported Cohen's $\kappa = 0.67$ for experienced physicians and $\kappa = 0.37$ for trainees, and confusion concentrated specifically between adjacent boundary classifications such as apparent equinus and crouch gait.
This intrinsic ambiguity motivates a deployment strategy that avoids hard classification for borderline cases that may be close to typically developing, referring them to 3D-IGA.
The 7-class Rodda and Graham classifier built on AGCN+ViT trial-level predictions presents poor hard-thresholded accuracy ($43\%$), but the macro-AUROC of $0.78 \pm 0.01$ confirms that the predicted z-scores retain substantial class-discriminative information well above chance for a seven-way problem.
Translating this aggregate signal into per-class trustworthiness combines \autoref{tab:rodda_per_class} and the confusion matrix in \autoref{fig:rodda_confusion}.
The most clinically trustworthy positive predictions are Jump (precision $0.73$, AUROC $0.76$), Crouch (precision $0.61$, AUROC $0.79$), and True Equinus (precision $0.46$, AUROC $0.84$), in that a clinician acting on a positive flag for one of these patterns is more likely than not to be correct.
Recurvatum is a different kind of result, in that the AUROC of $0.82$ shows the model can rank Recurvatum higher than non-Recurvatum trials yet recall of only $11\%$ means the model rarely commits to that flag, so the predicted Recurvatum z-score is useful as a continuous severity indicator while the hard label remains unreliable.
Apparent Equinus (precision $0.28$, AUROC $0.68$) and Ankle Crouch ($n = 10$, all metrics weak) are the least trustworthy classifications, with Apparent Equinus operating as a sink for ankle-underestimated trials as detailed above and Ankle Crouch lacking statistical power.
The practical implication is that the strongest single-camera video-based recommendations are positive flags for Crouch, Jump, and True Equinus, whereas Apparent Equinus flags should be reviewed against the ankle-region kinematics in the underlying video before clinical action.

\subsection{Generalization Across Diagnoses}

Unlike prior video-based gait studies that recruit single-diagnosis cohorts, our model was trained and evaluated on a heterogeneous tertiary clinical population spanning 60 primary diagnoses, with cerebral palsy the most common (35.8\%) and the remainder distributed across arthrogryposis, talipes equinovarus, spina bifida, and a long tail of rare neurological, musculoskeletal, and genetic conditions.
This breadth reflects the practical screening setting, where the underlying diagnosis is often unknown and the kinematic deviation itself, not the etiology, is what informs treatment decisions.
That Rodda and Graham z-scores can be regressed from monocular video across this diagnostic range suggests the predictor is anchored to underlying sagittal-plane kinematics rather than diagnosis-specific visual cues, supporting deployment as a diagnosis-agnostic kinematic screening tool.

\subsection{Limitations and Future Work}

First, although our cohort already spans 60 primary diagnoses all data originate from a single center (Shriners Children's Philadelphia), cross site evaluation is an important test of the systems generalizability to variation in scanner geometry, lighting, and patient population, which is currently unknown.
Second, the pose estimator was trained on normative gait and was not adapted to CP-specific movement kinematics, the resulting ankle and foot keypoint noise may be directly limiting ankle regression accuracy in particular.
Third, class imbalance is severe for Ankle Crouch ($n = 10$) and Recurvatum ($n = 29$), making per-class metrics unreliable for these classifications.
To address these limitations, we are currently working on cross site evaluation with Shriners Children's with data collected from Shriners Children's 14 integrated motion analysis Centers (MACs) across the country. 
There are several planned improvements to the pose estimation model used in the analysis, including fine-tuning for improved ankle prediction accuracy using synced mocap and video data of real pathological gaits, and modern backbones (ViT).

\section{Conclusion}

Cerebral palsy is the most common cause of life-long physical disability in childhood, and for the roughly three-quarters of children with CP who are ambulatory, preserving and optimizing walking function is the central goal of rehabilitation~\citep{rosenbaum_report_2007, oskoui_update_2013, armand_gait_2016}.
The gold standard for quantifying CP gait deviation, the Rodda and Graham classification derived from 3D-instrumented gait analysis, is locked behind specialized regional centers. The only widely available alternative, observational assessment, shows only moderate inter-rater agreement that drops further among less experienced clinicians~\citep{kim_reliability_2011, toro_review_2003}.
This work asked whether monocular clinical video alone, recorded during routine visits, can recover the Rodda and Graham knee and ankle z-scores that drive treatment decisions, evaluated across 152 children spanning 60 primary pediatric gait diagnoses.

Direct biomechanical computation of z-scores from monocular-derived joint angles is not viable. Every viewpoint yielded negative $R^2$, with even the best (sagittal) view performing worse than predicting the population mean, a limit of current single-view markerless pose estimation rather than of the z-score formulation itself.
A learned regressor closes this gap.
AGCN with ViT hierarchical aggregation~\citep{shi_two_2019, dosovitskiy_image_2021} reaches $R^2 = 0.80 \pm 0.02$ and CCC $= 0.89$ for knee z-scores and $R^2 = 0.57 \pm 0.02$ and CCC $= 0.72$ for ankle z-scores on the sagittal view.
Accuracy is strongly stratified by boundary proximity (easy CCC $= 0.89$ vs.\ hard CCC $= 0.27$), so regression accuracy alone is insufficient for the hard-edged seven-class Rodda and Graham assignment, which reaches $43 \pm 1\%$ accuracy with macro-AUROC $= 0.78 \pm 0.01$.
Binary screening for excess knee flexion is the system's strongest and most clinically deployable capability, achieving AUROC $= 0.88$ and recall $= 0.83$, sufficient to function as a video-based triage instrument that flags children for full 3D-IGA in settings where the laboratory itself is unavailable.
Ankle prediction remains the primary bottleneck across both regression and classification, traceable to a pose-estimator normative prior that systematically underestimates equinus and severe plantarflexion configurations rare in the typically developing motion-capture data on which the estimator was trained.
The cohort's 60-diagnosis breadth, with cerebral palsy the most common (35.8\%), further suggests that the predictor is anchored to underlying sagittal-plane kinematics rather than diagnosis-specific visual cues, supporting use as a diagnosis-agnostic kinematic screening tool.

Closing the remaining gap to clinical deployment will require fine-tuning the pose estimator on synced motion-capture and video data of real pathological gait to address the ankle bottleneck, and prospective multi-center validation with consumer-grade cameras in uncontrolled clinical environments to establish generalization beyond a single specialized center.
Even at present performance, video-based knee z-score estimation and binary excess-flexion screening offer a concrete path toward objective, scalable gait assessment for the children whose treatment decisions currently depend on either an unavailable laboratory or an unreliable visual judgment.

\section{Acknowledgements}
This study was funded by a grant (71013) from Shriners Children's. 
Hyeokhyen Kwon is partially funded by the National Institute on Deafness and Other Communication Disorders (grant No. 1R21DC021029-01A1), Georgia CTSA Pilot Grants Program, and Shriners Children's.

\bibliography{main}

\clearpage
\section*{Supplementary}

\paragraph*{Supplementary Table 1.}
\label{S1_Table}
{\bf Full primary-diagnosis breakdown of the 152-child cohort.} Counts and percentages for all 60 distinct primary diagnoses recorded at intake, ordered by frequency.

\begin{table}[!h]
\centering
\small
\begin{tabular}{lr@{\hspace{2em}}lr}
\toprule
\textbf{Diagnosis} & \textbf{n} & \textbf{Diagnosis} & \textbf{n} \\
\midrule
Cerebral Palsy & 54 & Hurler Syndrome & 1 \\
Arthrogryposis & 11 & Undiagnosed Genetic & 1 \\
Talipes Equinovarus & 10 & Undiagnosed & 1 \\
Spina Bifida & 5 & Left Hemiplegia & 1 \\
In-toeing & 4 & Flaccid Foot Drop & 1 \\
Toe Walker & 4 & Periventricular Leukomalacia & 1 \\
Femoral Anteversion & 3 & Weaver Syndrome & 1 \\
Spinal Cord Injury & 2 & Hemiparesis & 1 \\
Traumatic Brain Injury & 2 & Prader-Willi Syndrome & 1 \\
Aicardi Goutieres Syndrome & 2 & Escobar Syndrome & 1 \\
Gait Abnormality & 2 & Scoliosis & 1 \\
Pes Planus & 2 & Patellar Instability & 1 \\
Genetic (unspecified) & 2 & Patellar Tendonitis & 1 \\
Ehlers-Danlos Syndrome & 2 & Sever's Disease & 1 \\
Asperger's Syndrome & 1 & Marfan Syndrome & 1 \\
Fibular Hemimelia & 1 & Autism Spectrum Disorder & 1 \\
Lower Extremity Neuropathy & 1 & Sacral Agenesis & 1 \\
Pierre-Robin Syndrome & 1 & Larsen Syndrome & 1 \\
Multiple Epiphyseal Dysplasia & 1 & Femoral Retroversion & 1 \\
Limb-Girdle Muscular Dystrophy & 1 & Genu Valgum & 1 \\
Bilateral Foot Pain & 1 & Blount's Disease & 1 \\
Congenital Hip Dysplasia & 1 & Acute Flaccid Myelitis & 1 \\
Traumatic Amputation & 1 & Familial Spastic Paraparesis & 1 \\
Leg Length Discrepancy & 1 & Autism & 1 \\
Internal Capsule Brain Tumor & 1 & KIF1A-related Disorder & 1 \\
Chromosome 16p11.2 Deletion & 1 & Joint Hypermobility & 1 \\
Knee Injury & 1 & Ellis-van Creveld Syndrome & 1 \\
Vein of Galen Malformation & 1 & Heel Cord Contractures & 1 \\
Nance-Horan Syndrome & 1 & Shaken Baby Syndrome & 1 \\
ACL Tear & 1 & Witteveen-Kolk Syndrome & 1 \\
\midrule
\multicolumn{4}{l}{\textbf{Total: 152 participants across 60 distinct primary diagnoses.}} \\
\bottomrule
\end{tabular}
\end{table}

\end{document}